\documentclass[letterpaper, 10 pt, conference]{ieeeconf}  % Comment this line out if you need a4paper

\IEEEoverridecommandlockouts                              % This command is only needed if 
\usepackage{amsmath}
\usepackage{tabularx}
\usepackage{amssymb}
\usepackage{graphics}
\usepackage{graphicx}
\usepackage{multirow}

\usepackage{url}
\usepackage{xparse}
\usepackage{xspace}
\usepackage{nicefrac}
\usepackage{microtype}
\usepackage{amsfonts}
\usepackage{enumerate}
\usepackage{setspace}
\usepackage{wrapfig}
\usepackage{subfigure}
\usepackage{booktabs} %
\usepackage{tikz}
\usepackage{hyperref}
\usepackage[style=ieee, backend=bibtex]{biblatex}
\usepackage{amsbsy}
\usepackage[export]{adjustbox}
\title{\LARGE \bf
Composing Pre-Trained Object-Centric Representations for Robotics From ``What'' and ``Where'' Foundation Models
\\[-1.0ex]
}

\author{Junyao Shi$^{*1}$, Jianing Qian$^{*1}$, Yecheng Jason Ma$^{1}$, Dinesh Jayaraman$^{1}$% <-this % stops a space
\\
\href{https://sites.google.com/view/pocr}{sites.google.com/view/pocr}
\\
\thanks{$^*$Equal Contribution.}%
\thanks{$^{1}$Computer and Information Science, University of Pennsylvania. Email: \texttt{junys, jianingq, jasonyma, dineshj@seas.upenn.edu}}%
\\[-8.0ex]
}

\newcommand{\methodname}{POCR\xspace}

\newcommand{\revise}[1]{\textcolor{black}{#1}}

\newcommand{\revisetwo}[1]{\textcolor{black}{#1}}
\usepackage{soul}

\usepackage[colorinlistoftodos]{todonotes}
\begin{document}

\maketitle
\thispagestyle{empty}
\pagestyle{empty}

%%%%%%%%%%%%%%%%%%%%%%%%%%%%%%%%%%%%%%%%%%%%%%%%%%%%%%%%%%%%%%%%%%%%%%%%%%%%%%%%
\begin{abstract}

There have recently been large advances both in pre-training visual representations for robotic control and segmenting unknown category objects in general images. To leverage these for improved robot learning, we propose POCR, a new framework for building pre-trained object-centric representations for robotic control. Building on theories of ``what-where'' representations in psychology and computer vision, we use segmentations from a pre-trained model to stably locate across timesteps, various entities in the scene, capturing ``where'' information. To each such segmented entity, we apply other pre-trained models that build vector descriptions suitable for robotic control tasks, thus capturing ``what'' the entity is. Thus, our pre-trained object-centric representations for control are constructed by appropriately combining the outputs of off-the-shelf pre-trained models, with no new training. On various simulated and real robotic tasks, we show that imitation policies for robotic manipulators trained on \methodname achieve better performance and systematic generalization than state of the art pre-trained representations for robotics, as well as prior object-centric representations that are typically trained from scratch. 
\end{abstract}

%%%%%%%%%%%%%%%%%%%%%%%%%%%%%%%%%%%%%%%%%%%%%%%%%%%%%%%%%%%%%%%%%%%%%%%%%%%%%%%%
\section{Introduction}

One of the fundamental challenges of developing general-purpose robots is how to build informative and generalizable visual representations that permit robots to acquire diverse manipulation skills. Pre-trained unsupervised vector representation encoders have matured and are fast becoming the de facto standard descriptors of the contents of raw sensory inputs for downstream tasks in language~\cite{brown2020language,devlin2018bert,radford2021learning,girdhar2023imagebind} and vision~\cite{radford2021learning,he2022masked,girdhar2023imagebind}. 
Recent works~\cite{xiao2022masked,nair2022learning,ma2022vip,ma2023liv} have also shown that the same paradigm could be applied to robotics by leveraging internet images~\cite{deng2009imagenet} and human videos~\cite{grauman2022ego4d}. Pre-trained representations for robotic control have clear advantages over representations trained in-domain: they can be flexibly deployed off-the-shelf, and they do not require large amounts of expensive task-specific data for learning.

While these pre-trained control-aware encoders have been proven useful for policy learning, they don't explicitly capture discrete and meaningful entities such as objects, which are essential for understanding the observation and reasoning about actions. In humans, per the “what-where” representation theory~\cite{goodale1992separate, ungerleider1994and, de2011usefulness} in cognitive science, the brain uses specialized neural pathways to encode two types of information: “what” information, which refers to the identity, features, and properties of an entity; and “where” information, which refers to the location, direction, and distance of an entity. A growing literature on object-centric embeddings (OCEs)~\cite{Locatello2020ObjectCentricLW,Burgess2019MONetUS,Engelcke2021GENESISV2IU,Lin2020SPACEUO,Sauvalle2022UnsupervisedMS,Kipf2019ContrastiveLO,Devin2017DeepOR,Wang2018DeepOP,Kumar2022GraphIRL,Zhu2022VIOLA,Sieb2019GraphStructuredVI,Kulkarni2019UnsupervisedLO,Minderer2019UnsupervisedLO} instantiates these ideas in artificial intelligence, commonly focusing on co-training the ``what'' and ``where'' pathways within a target domain. 

We investigate a simpler route to OCEs suitable for robots, paved by: (1) recent advances in image segmentation~\cite{Kirillov2023SegmentA,Zou2023SegmentEE}, the task of identifying groups of pixels in an image that correspond to semantic objects and their parts. These pre-trained models can now reliably locate the discrete entities in in-the-wild images in arbitrary domains, and (2) recently proposed pre-trained representations for control.

\begin{figure*}
\centering 
% \vspace{-0.2in}
\includegraphics[width=0.9\textwidth]{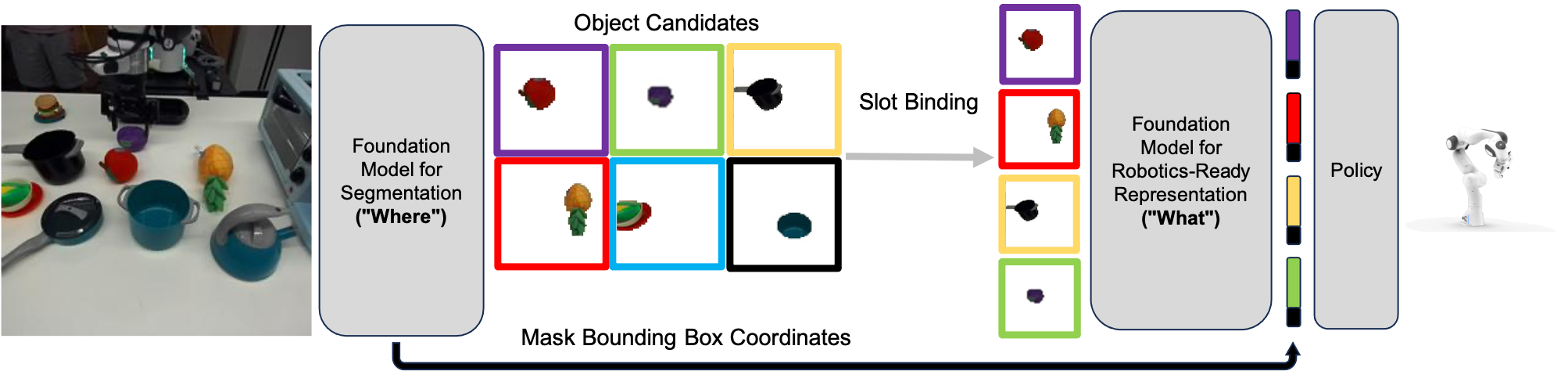}
\vspace{-0.1in}
\caption{\textbf{\methodname}: \textbf{P}re-Trained \textbf{O}bject-\textbf{C}entric \textbf{R}epresentations for Robotics by chaining ``what'' and ``where'' foundation models. The ``where" foundation model produces a set of masks representing objects in the scene. Slot binding selects which among them to bind to the slots in our OCE. Image contents in each slot are represented by the ``what" foundation model and their mask bounding box coordinates. The robot learns policies over slot representations.}
\vspace{-0.2in}
\label{figure:concept-figure}
\end{figure*}

We propose pre-trained object-centric representations for robotics (\methodname), a general-purpose ready-made model constructed by appropriately chaining segmentation and control-aware vision ``foundation models''. Each such model has individually been pre-trained on large and diverse datasets, and afterwards been shown to work well on many domains of interest. A composite representation that inherits these generalization properties may be used off-the-shelf in arbitrary new robotics tasks; see Figure~\ref{figure:concept-figure} for a schematic overview.

In our experiments, we study various choices of foundation models to plug-and-play in the \methodname framework. 
On unseen simulated and \textit{real-world} robotic manipulation tasks, we find that the best \methodname representations enable significantly better policy learning than all prior representations. 
\methodname policies even demonstrate systematic generalization to unseen test-time variations in the scene. In summary, our findings suggest that \methodname provides a simple framework for generating, to our knowledge, the very first generic pre-trained object-centric representations for robotics that can reliably be used off-the-shelf in new robotic environments and tasks.

\section{Problem Setup and Background}\label{sec:bg}
We are interested in sample-efficient learning of robotic manipulation policies in arbitrary multi-object scenes, with possible distractor objects. For example, in our real robot experiments, we task a robot arm attached to a cluttered kitchen countertop with moving fruits and vegetables into pots and pans around it, with only a few tens of demonstrations.

\textbf{Pre-Trained Representations for Control.} Our work builds on top of the literature on pre-trained visual representations for control~\cite{nair2022r3m,ma2022vip,ma2023liv,xiao2022masked,radosavovic2023real,majumdar2023we}.
These works have shown how frozen visual representations, pre-trained on out-of-domain data, can serve as effective visual encoders for policy learning on unseen robot tasks. However, flat image-level representations typically lose fine-grained object-centric information often necessary for solving tasks that require reasoning about multi-object relationships, object affordance, or object articulations~\cite{heravi2022visuomotor}.  
Indeed, in our experiments, we discovered that they struggle with learning actions in multi-object
scenes.
Nevertheless, they serve as effective ``what" descriptors in our compositional ``what-where" framework.

\textbf{Object-Centric Embeddings (OCEs), Their Pros and Cons.} We propose to enable manipulation tasks with OCEs of visual scenes. Many prior works~\cite{Locatello2020ObjectCentricLW,Burgess2019MONetUS,Engelcke2021GENESISV2IU,Lin2020SPACEUO,Sauvalle2022UnsupervisedMS, Kipf2019ContrastiveLO} formulate OCEs 
using two components: (1) the location, or ``where'' component, $l_i$ indicates the presence and location of an entity, such as through a segmentation mask~\cite{Locatello2020ObjectCentricLW,Burgess2019MONetUS,Engelcke2021GENESISV2IU,Lin2020SPACEUO,Sauvalle2022UnsupervisedMS}, bounding box~\cite{Devin2017DeepOR,Wang2018DeepOP,Kumar2022GraphIRL,Zhu2022VIOLA,Sieb2019GraphStructuredVI}, or keypoint location~\cite{Kulkarni2019UnsupervisedLO,Minderer2019UnsupervisedLO}, and (2) the content, or ``what'' component, captures the properties of the object such as its texture, pose, and affordances. 

OCEs disentangle scene objects, enabling improved systematic generalization, symbolic reasoning, sample-efficient learning, and causal inference starting from visual inputs~\cite{van2019perspective, Greff2020-hq, Lake2016-ze, pmlr-v162-dittadi22a,yoon2023investigation} compared to unstructured ``flat'' vector representations of the scene. They can also serve as a shared representation interface \cite{SAVi} between humans and robots, which is potentially useful for task specification. 

Despite all these potential functional advantages of OCEs, state-of-the-art pre-trained approaches in robot learning today commonly use flat vector representations of the scene~\cite{Haldar2023TeachAR,Young2021PlayfulIF, nair2022r3m, xiao2022robotic, ma2022vip}. We argue that this is primarily because training deep neural networks to generate OCEs is difficult; they require non-standard architectures and do not train as stably. This in turn means that current deep OCEs are restricted to be relatively small-capacity networks that are highly sensitive to architectural choices~\cite{Engelcke2020-rz,papa2022inductive}. They must therefore be trained on domain-specific data, and even then, on large image datasets in relatively small domains. Leave alone re-using pre-trained OCEs in new task domains, state-of-the-art OCEs perform poorly even in-domain in realistic, visually complex settings~\cite{yang2022promising}, as we also find in our experiments.

Thus today's OCEs trail flat representations in practical utility, whereas flat representations trail OCEs in functionality. 
Our attempt aims to get the best of both worlds by building functional and practical pre-trained OCEs for robotics. 

\section{Pre-Trained Object Centric Representations For Robotic Manipulation}

We target an OCE that at each time $t$ summarizes the scene $o^t$ in terms of discrete ``slots'' $s^t_i$, that correspond to the entities in the scene, i.e., objects and parts. To unclutter notation, we will omit the time index $t$ when it is not relevant. Each slot is a tuple $s_i=(l_i, z_i)$ with two components: (1) the location $l_i$ and (2) the content, often called a ``slot vector'' $z_i \in \mathbb{R}^D$, visible in the scene regions $o[l_i]$ identified by $l_i$. Since segmentations provide fine-grained information about object entities, they are an ideal choice for representing $l_i$. However, prior methods that use masks inevitably require extensive in-domain training, since no pre-trained models are designed to encode masks. Meanwhile, pre-trained vision encoders can be used off-the-shelf for representing image content, but their flat representations don't explicitly capture object entities. It is thus a priori non-obvious how to properly combine masks and pre-trained encoders to properly represent $l_i$ and $z_i$ in an OCE framework. Towards our unique approach to chain ``where'' and ``what'' foundation models into a useful representation for manipulation policy learning, we address three key questions: \textbf{where} are the object regions, \textbf{what} are their contents, and \textbf{how} should robots 
act to accomplish manipulation tasks given such what-where representations. 

\subsection{The Where: Localizing and Assigning Objects to Slots}\label{sec:pocr-the-where}

To go from segmentation outputs to 
slot masks $l_i$, we propose a lightweight procedure to match each mask to object segments in a reference image (e.g., the initial frame).

\textbf{Obtaining object segments in a reference image by screening the object-level foreground entity candidates.} 
We collect some reference images, such as from the initial frames of demonstrations, and we use them to compute the background regions in the image following the procedure in~\cite{Amir2021DeepVF}. 
Then, on one of these same reference images $o^{ref}$, we run the segmentation model to produce the set of masks $m_i^{ref}$. By design, a general-purpose segmentation model produces 
an over-complete set of segmentation masks corresponding to various entities $i$ in the scene at various levels of granularity, 
and also irrelevant entities in background regions of the scene.  

To discard background and incomplete entities among these reference image masks, we identify object-level foreground entities among $m_i^{ref}$ as follows. We use a greedy non-maximum suppression algorithm: sorting the masks in decreasing order of foreground area, we iteratively select masks $m_i$ that do not overlap with either previously selected masks or the background regions. The end result is a much shorter list of $n$ mask candidates $\{l_1^{ref}, \ldots, l_k^{ref}\}$ for object slot binding in the next step. The number of slots $k$ is set to a large constant independent of the task or frame, and if segmentation proposals are fewer than $k$, the extra slots are treated as empty.

\textbf{Localizing objects in each observation via consistent slot binding.} Given these selected reference masks that encode objects, the slots in our desired OCE must bind to these objects in each image. Towards this, we first apply the aforementioned procedure to screen the object-level foreground entity candidates to obtain a shorter list of mask candidates $\{m_1(o), \ldots, m_n(o)\}$ in image $o$. Next, to decide which among them to bind to the $k$ slots in our OCE of the image $o$, we perform a Hungarian matching~\cite{Kuhn1955TheHM} between the $n$ selected candidates $\{m_1(o), \ldots, m_n(o)\}$, and the $k$ masks  $\{l_1^{ref}, \ldots, l_k^{ref}\}$ representing object slots in the reference image. For the matching costs, we compute the cosine distance between pre-trained DINO-v2 representations of each slot mask, obtained through ROI-pooling. The final output is an ordered set of $k$ masks $\{l_1(o), \ldots, l_k(o)\}$. In our method, each mask is represented by its bounding box coordinates, which serve as the ``where'' component of our OCE.

\subsection{The What: Representing The Image Contents in Each Slot}\label{sec:pocr-the-what}

Given slot masks $\{l_{1}(o),\dots,l_{k}(o)\}$ for image $o$, we must compute, for each slot, its ``slot vector'' $z_i$. This slot vector captures the properties of the object visible in the scene regions specified by $l_i$, i.e., ``what'' is at $l_i$? As foreshadowed above, we will use pre-trained vision encoders to compute these slot vectors. For each slot $i$, we first generate a corresponding masked RGB image $o_i$ by element-wise multiplying the binary mask $l_i$ with the image $o$, and then compute image representations $z_i = \text{encoder}(o_i)$ over it. Together, $s(o)=\{(z_i(o),l_i(o))\}_{i=1}^k$ 
constitutes \methodname, our ``plug-and-play'' OCE framework for robotics.

\subsection{The How: Learning Robot Manipulation Policies from Demonstrations with \methodname}
\label{sec:atten}
So far, no learning has occurred as we have leveraged pre-trained models to format visual observations into OCEs. 
Given that the object binding operation may be sensitive to noise and occasionally make incorrect assignments, it is natural to use policy architectures that encode permutation invariance~\cite{zaheer2017deep,vaswani2017attention,kipf2016semi, velivckovic2017graph,zhou2022policy}.
We employ a self-attention (SA)~\cite{vaswani2017attention} neural layer to process the OCEs, and then aggregate the outputs to feed into an MLP policy. 
\begin{equation}
\pi_{\theta}(\{(z_i, l_i)\}_{i=1}^{k})) := \text{MLP}_{\theta}\left(\sum_{i=1}^{k}\text{SA}_{\theta}[z_i, l_i]\right),
\end{equation}
where $[\cdot,\cdot]$ is the concatenation operation. We train policies with a mean squared error (MSE) behavior cloning loss to predict the expert actions in the provided demonstrations. 

\section{Other Related Work}\label{sec:other-related}
\textbf{Traditional Uses Of Object Detectors In Robotics.} 
In a way, our approach of combining pre-trained models into one OCE encoder with no training is reminiscent of more traditional and modular approaches to representing visual scenes in robotics, such as by computing hand-defined (e.g., SIFT, HOG) features over object detector outputs~\cite{lowe2004distinctive, dalal2005histograms}. Such approaches have continued to be useful since the advent of deep learning, e.g., recent works have employed detectors for object poses~\cite{Tremblay2018DeepOP,Ye2019ObjectcentricFM,Migimatsu2019ObjectCentricTA, zhu2021hierarchical} and bounding boxes~\cite{Devin2017DeepOR,Wang2018DeepOP,Kumar2022GraphIRL,Zhu2022VIOLA,Sieb2019GraphStructuredVI, mao2023pdsketch, jiang2022vima}. Given the abundance of research in object detection from the computer vision community, those works either assume ground-truth object states~\cite{silver2022learning}, leverage existing object detectors~\cite{Wang2018DeepOP,Kumar2022GraphIRL,Sieb2019GraphStructuredVI, mao2023pdsketch, jiang2022vima} such as Mask R-CNN~\cite{MaskRCNN} or incorporate vision backbones such as a region proposal network~\cite{Ren2015FasterRT} for general object proposals and then train a policy that attends to the task-relevant information~\cite{Devin2017DeepOR,Zhu2022VIOLA}.  However, these methods typically require prior knowledge of object categories thus failing to handle previously unseen objects. These methods are also data-intensive, requiring significantly larger in-domain task-specific datasets than ours to learn or refine the OCE. Indeed, the growing literature on unsupervised deep OCEs is motivated by moving beyond such domain-specific labeled datasets, but has its own disadvantages, as we motivated in Sec~\ref{sec:bg}. 

\textbf{Concurrent Works That Use Foundation Models for Segmentation.} Concurrent works~\cite{ferraro2023focus, yang2023pave, zhu2023learning, garg2023robohop} also study the usage of general-purpose segmentation models such as SAM~\cite{Kirillov2023SegmentA} for control. FOCUS~\cite{ferraro2023focus} uses SAM to generate mask supervision for a model-based agent that learns an object-centric world model. MMPM~\cite{yang2023pave} employs SAM to obtain language-grounded object masks given their bounding boxes. GROOT~\cite{zhu2023learning} utilizes SAM to construct object-centric 3D representations. RoboHop~\cite{garg2023robohop} leverages SAM to generate topological segment-based map representation for robot navigation. Among them,  only GROOT~\cite{zhu2023learning} and RoboHop~\cite{garg2023robohop} handle object tracking in complex scenes, and none of them propose a viable pre-trained OCE for robotics. 

We set up baselines that are similar in spirit to GROOT and RoboHop: ``SAM-Scratch'', like GROOT and unlike \methodname, embeds SAM masks with an encoder trained from scratch along with the task policy. ``SAM-centroid'', like RoboHop, uses segment centroids to represent each object segment. 

We develop \methodname to be the first reusable OCE that can be applied ``off the shelf'' to a large range of robotic environments and tasks. Concurrent with this work, SOFT~\cite{qian2024soft} shows how OCEs can be constructed even from pre-trained vision models that do not directly output object segments.

\section{Experimental Results}
\label{sec:result}

Our experiments aim to answer the following questions: 1) What is the appropriate ``where" representation for \methodname? 2) What is the appropriate ``what" representation for \methodname 3) Does \methodname enable better policy learning compared to prior pre-trained representations or object-centric methods for robotics? 4) Does \methodname enable systematic generalization? Video results and supplementary materials: \href{https://sites.google.com/view/pocr}{sites.google.com/view/pocr}. 

\subsection{Simulation Experiments Setup}
\label{sec:sim-env-method}
\begin{figure}
\resizebox{.5\textwidth}{!}{% 
\subfigure[Pick up Cup]{\label{fig:pickupcup}\includegraphics[width=0.12\textwidth]{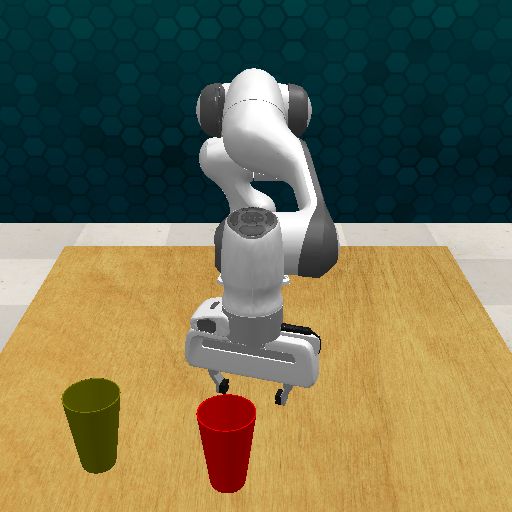}}
\subfigure[Rubbish in Bin]{\label{fig:putrubbishinbin}\includegraphics[width=0.12\textwidth]{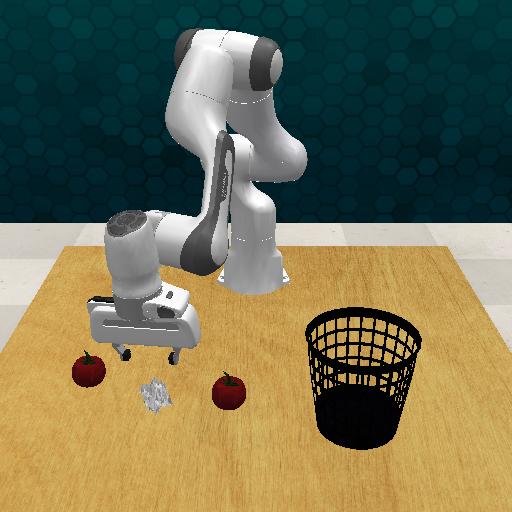}}
\subfigure[Stack Wine]{\label{fig:stackwine}\includegraphics[width=0.12\textwidth]{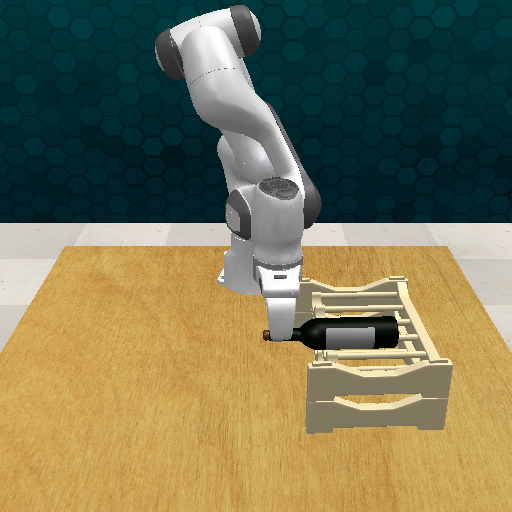}}
\subfigure[Phone on Base]{\label{fig:phoneonbase}\includegraphics[width=0.12\textwidth]{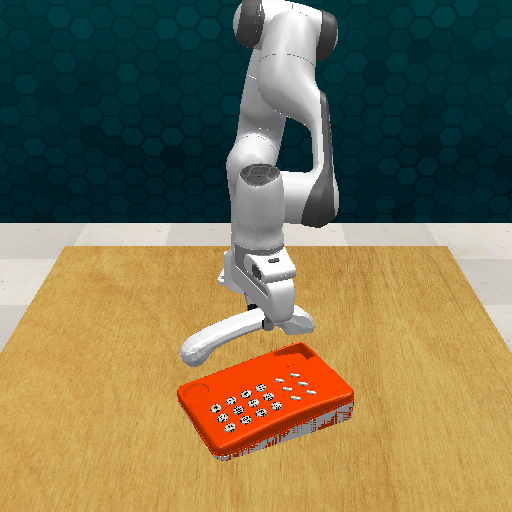}}
}
\\
\resizebox{.5\textwidth}{!}{% 
\subfigure[Water Plants]{\label{fig:waterplants}\includegraphics[width=0.12\textwidth]{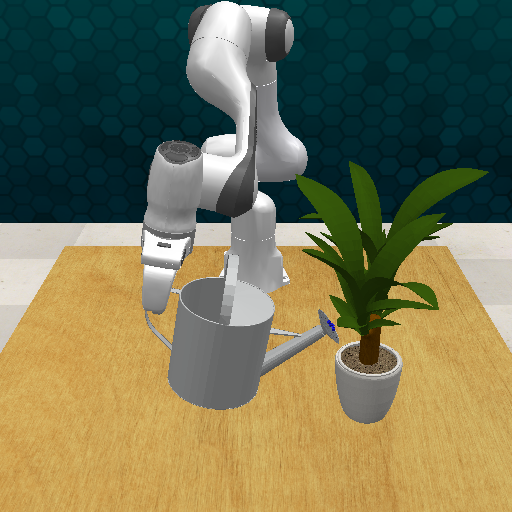}}
\subfigure[Close Box]{\label{fig:closebox}\includegraphics[width=0.12\textwidth]{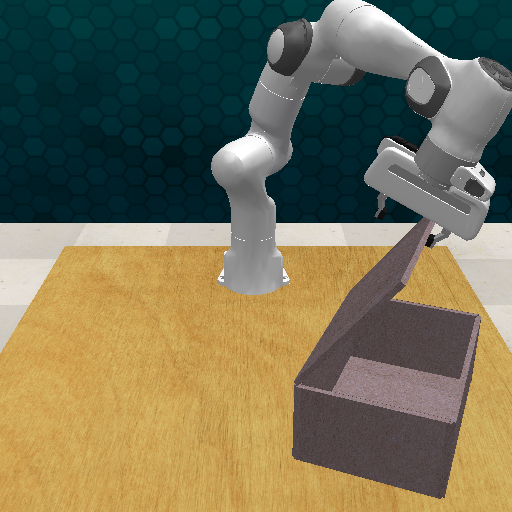}}
\subfigure[Close Laptop]{\label{fig:closelaptop}\includegraphics[width=0.12\textwidth]{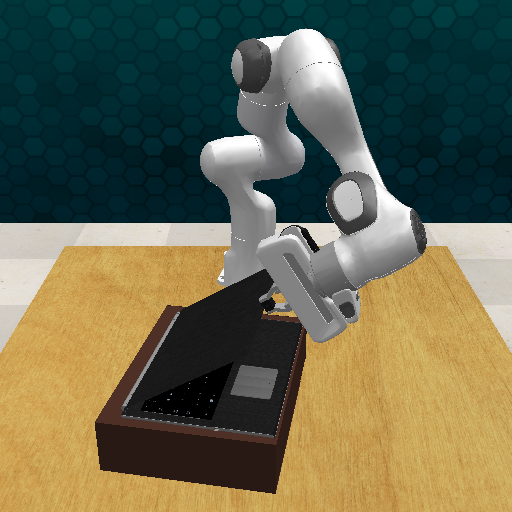}}
\subfigure[Real Robot]{\label{fig:realworld}\includegraphics[width=0.12\textwidth]{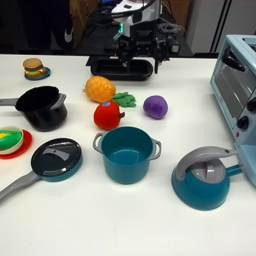}}
}
\vspace{-0.15in}
\caption{Evaluation Environments.}
\label{figure:environments}
\vspace{-0.25in}
\end{figure}

\textbf{Environments.} We selected 7 tasks (Figure~\ref{figure:environments}) in RLBench~\cite{james2020rlbench} as our simulation testbed to validate our algorithmic design. 
\texttt{Pick up Cup} and \texttt{Put Rubbish in Bin} are multi-object tasks with distractor objects. \texttt{Stack Wine}, \texttt{Phone on Base}, \texttt{Water Plants} are multi-object tasks that require reasoning about the geometry and affordance of objects to manipulate them into a desirable configuration. \texttt{Close Box} and \texttt{Close Laptop} are single-object tasks that require reasoning about object articulations. The poses of all objects are randomized for each episode. In \texttt{Pick up Cup}, the color of the distractor cup is also randomized.

\textbf{Policy Learning.} Our policy training and evaluation protocol mostly follows~\cite{james2022q}; in particular, for each task, we use 100 demonstrations collected using a state-based motion planner and train single-task policies using behavior cloning. The action space is Franka robot's 6-DOF end-effector pose and gripper state, and we use keyframe action representation to reduce the task horizon. 
For each method, we train policies using 3 seeds and report the mean and the standard error of the maximum success rates each seed achieves during training on 100 evaluation rollouts, following standard practice~\cite{nair2022r3m}. See \href{https://sites.google.com/view/pocr}{Supplementary Materials} I-A for more experimental details.

\subsection{Investigating ``Where" Representations for \methodname}
\label{sec:where-comparison}
\begin{figure}
\centering
\includegraphics[width=0.18\columnwidth]{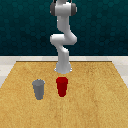}
\includegraphics[width=0.18\columnwidth]{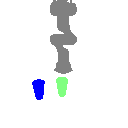}
\includegraphics[width=0.18\columnwidth]{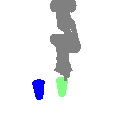}
\includegraphics[width=0.18\columnwidth]{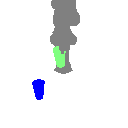} \\
\hspace{0.04cm}
\includegraphics[width=0.18\columnwidth]{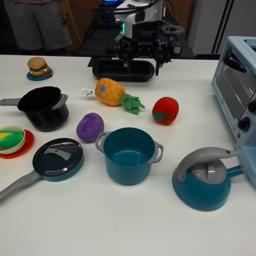}
\includegraphics[width=0.18\columnwidth]{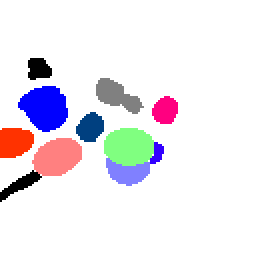}
\includegraphics[width=0.18\columnwidth]{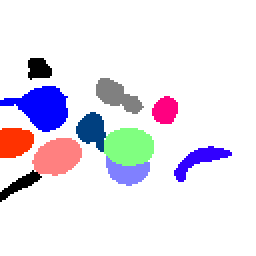}
\includegraphics[width=0.18\columnwidth]{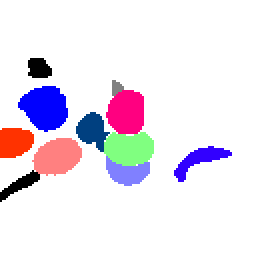}
\vspace{-0.1in}
\caption{\methodname segmentation results over demonstrations.}
\label{figure:qualitative}
\vspace{-0.1in}
\end{figure}

\methodname's ``where'' representation involves Segment Anything Model (SAM)~\cite{Kirillov2023SegmentA}. We first investigate the benefits of using SAM over prior approaches that train such segmenters in-domain. 
We compare SAM to the best such approach simulation: AST-SEG~\cite{Sauvalle2022UnsupervisedMS}, an unsupervised method, and SAM~\cite{Kirillov2023SegmentA}, a pre-trained method. We train AST-SEG on our demonstrations in RLBench (about 1400 images for \texttt{Pick up Cup} and 2500 images for \texttt{Rubbish in Bin}) and use SAM directly out-of-the-box. We report the quantitative results with foreground adjusted random index (FG ARI)~\cite{Rand1971ObjectiveCF, Hubert1985ComparingP}, a standard segmentation metric. SAM achieves 0.99 FG ARI scores (max is 1) on both tasks, while AST-SEG fails to segment almost all the foreground objects, scoring only 0.2 on \texttt{Pick up Cup} and 0.1 on \texttt{Rubbish in Bin}. We reason that AST-SEG and prior segmentation methods~\cite{girshick2015fast, wang2017fast} typically require large domain-specific datasets for unsupervised training. This is unsuitable for sample-efficient policy learning in a new environment. Therefore, we employ SAM as the ``where" representation for \methodname. Figure~\ref{figure:qualitative} shows the visualizations of SAM masks after slot binding in various environments, illustrating that integrating SAM into our pipeline provides consistent and accurate tracking of objects. See 
\href{https://sites.google.com/view/pocr}{Supplementary Materials} III for additional visualizations.

In \href{https://sites.google.com/view/pocr}{Supplementary Materials} II-A, we show that representing \methodname's ``where" component explicitly using SAM mask bounding box coordinates (\textbf{SAM-bbox}) leads to slightly better downstream control results than using SAM mask centroids (\textbf{SAM-centroid}). It also leads to significantly better results than using no explicit ``where" representation, suggesting that the ``where" information is not well captured implicitly in the ``what" encodings. In \href{https://sites.google.com/view/pocr}{Supplementary Materials} II-D, we show that \methodname's object binding procedure has near-perfect accuracy when evaluated quantitatively using \textit{ground-truth} masks.
However, as presented in \href{https://sites.google.com/view/pocr}{Supplementary Materials} II-B, there still exists a small gap between policies trained with SAM masks and \textit{ground-truth} masks. 

\begin{table}[ht]
\centering
\caption{Behavioral cloning success rates (avg. over 100 rollouts), ablation of various ``What" Representations for \methodname.}
\vspace{-0.1in}
\resizebox{.5\textwidth}{!}{% <------ Don't forget this %
\begin{tabularx}{1.2\columnwidth}{p{3.6cm}|p{2.6cm}p{3.0cm}@{}}
\toprule
``What'' Representations & \texttt{Pick up Cup} & \texttt{Rubbish in Bin} \\
\midrule
\pmb{SAM-LIV} & \pmb{79.3 $\pm$ 0.7} & \pmb{35.3 $\pm$ 1.8} \\
SAM-R3M & 78.0 $\pm$ 1.3 & 33.3 $\pm$ 1.3 \\
SAM-ImageNet & 70.7 $\pm$ 0.7 & 30.0 $\pm$ 2.0 \\
\midrule
SAM-Scratch & 32.7 $\pm$ 2.9  & 12.0 $\pm$ 2.0 \\
\midrule
SAM-None & 55.7 $\pm$ 1.8 & 24.6 $\pm$ 1.3  \\
\bottomrule
\end{tabularx}
}
\label{table:what-reps}
\vspace{-0.15in}
\end{table}

\begin{table*}[!htbp]
\caption{Behavioral cloning success rates (avg. over 100 rollouts), comparing POCR with Prior Representations. }
\vspace{-0.1in}
\centering
\resizebox{1.0\textwidth}{!}{% <------ Don't forget this %
\begin{tabularx}{1.4\textwidth}{p{3.0cm}|p{3.0cm}@{}@{}p{3.0cm}@{}|p{3.0cm}@{}@{}p{3.0cm}@{}@{}p{3.0cm}@{}|p{3.0cm}@{}@{}p{3.0cm}}
\toprule
\multicolumn{1}{p{3.0cm}|}{} & \multicolumn{2}{c|}{Multi-Object with Distractors} & \multicolumn{3}{c|}{Multi-Object with Geometric Reasoning} & \multicolumn{2}{c}{Articulated Objects}\\
\midrule
 Representations & Pick up Cup & Rubbish in Bin & Stack Wine & Phone on Base & Water Plants & Close Box  &  Close Laptop \\
\midrule
\pmb{\methodname(SAM-LIV)} & \pmb{79.3 $\pm$ 0.7} & \pmb{35.3 $\pm$ 1.8} & \pmb{54.0 $\pm$ 2.3} & \pmb{34.7 $\pm$ 3.3} & \pmb{16.3 $\pm$ 0.9} & \pmb{96.7 $\pm$ 0.7} & 32.7 $\pm$ 1.3 \\ \midrule
LIV & 49.3 $\pm$ 1.9 & 21.9 $\pm$ 0.8 & 24.0 $\pm$ 5.0 & 18.7 $\pm$ 2.4 & 4.0 $\pm$ 1.2 & 86.0 $\pm$ 5.3 & \pmb{35.3 $\pm$ 1.8} \\
R3M & 52.3 $\pm$ 1.5 & 20.7 $\pm$ 1.8 & 40.0 $\pm$ 2.0 & 27.3 $\pm$ 2.4 & 8.0 $\pm$ 2.0 & 85.3 $\pm$ 1.3 & 31.3 $\pm$ 1.3 \\
ImageNet & 14.0 $\pm$ 1.2 & 6.7 $\pm$ 0.7 & 11.3 $\pm$ 2.4 & 7.3 $\pm$ 1.3 & 7.3 $\pm$ 0.7 & 58.7 $\pm$ 1.3 & 30.0 $\pm$ 2.0 \\
\midrule
VIMA (GT masks) & 31.0 $\pm$ 6.7 & 5.7 $\pm$ 4.0 & N/A & N/A & N/A & N/A & N/A \\
\midrule
Scratch & 15.9 $\pm$ 1.0 & 2.7 $\pm$ 0.4 & 1.3 $\pm$ 0.7 & 3.3 $\pm$ 0.7 & 2.0 $\pm$ 0.0 & 86.7 $\pm$ 1.8 & 30.0 $\pm$ 2.3 \\
\bottomrule
\end{tabularx}
}
\label{table:baselines}
\end{table*}

\subsection{Investigating ``What" Representations for \methodname}
\label{sec:what-comparison}
\textbf{Representations.} Given that SAM masks, represented as the bounding box coordinates of each mask, provide the best ``where" representation for \methodname, we thoroughly investigate which ``what" representation provides the best control performance by ablating various pre-trained vision encoders for control in simulation. \textbf{LIV}~\cite{ma2023liv} and \textbf{R3M}~\cite{nair2022r3m} are visual representations for robot control pre-trained on large-scale in-the-wild human videos. \textbf{ImageNet} refers to a ResNet-50 network~\cite{he2016deep} pre-trained on the ImageNet dataset~\cite{deng2009imagenet}. We also consider training a CNN network from scratch as the visual encoder of the masks (\textbf{Scratch}), using the official implementation from~\cite{james2022q}.
Finally, we consider using only the bounding box of SAM masks 
without additional ``what" slot vector representations (\textbf{None}).

\textbf{Results.} As shown in Table~\ref{table:what-reps}, when combined with SAM as the ``what" representation in \methodname, LIV performs best.
Training from scratch (SAM-Scratch) performs significantly worse than all other ``what" encoders, which validates the advantage of using pre-trained foundation models over in-domain training. Among the pre-trained ``what'' encoders, LIV yields a small but consistent advantage over R3M and ImageNet. We retain LIV as the ``what'' encoder of choice for all following experiments.

\subsection{How Does \methodname Compare to SOTA Representations?}
\label{sec:simulation-experiments}
To our knowledge, \methodname representations are the first generic pre-trained object-centric representation for robotics. We now evaluate them against state-of-the-art pre-trained representations and prior object-centric methods.

\textbf{Baselines.} Our most relevant baselines are prior pre-trained methods, especially those for robotics. We again compare to \textbf{LIV}, \textbf{R3M}, and \textbf{ImageNet}. We also compare to \textbf{VIMA}~\cite{jiang2022vima}, a state of the art object-centric baseline that parses images into object bounding boxes and learns transformer-based object representations from scratch. Instead of following VIMA's setup to train an object detector with tens of thousands of in-domain mask annotations, we supply VIMA with ground-truth masks generated by RLBench. In spirit, \textbf{VIMA} is similar to \textbf{SAM-Scratch} in Table~\ref{table:what-reps}. Finally, we again consider training a CNN network from scratch (\textbf{Scratch}) as the visual descriptor of the image observation.

\textbf{Simulation Results.} Table~\ref{table:baselines} shows the results. 
Each flat scene-level pre-trained representation (LIV, R3M, ImageNet) trails behind its POCR counterpart (SAM-LIV, SAM-R3M, SAM-ImageNet), reported in Table~\ref{table:what-reps}, showing the versatility of the POCR framework. Naturally, as the best POCR, SAM-LIV far outperforms these flat encoders. Next, in our experiments, the SOTA object-centric approach VIMA incurred high compute and time costs for training, so we only trained it on 2 tasks, but the conclusions are clear: VIMA, which only pre-trains its object detectors and trains ``what'' representations from scratch performs poorly in this limited-demonstration setting.
\methodname with SAM for ``where'', and LIV for ``what'' thus outperforms the best current representations for multi-object manipulation. In \href{https://sites.google.com/view/pocr}{Supplementary Materials} II-C, we show that prior methods also fall far short of \methodname when used as representations in BC-RNN~\cite{mandlekar2021matters}, a popular approach for recurrent imitation policy learning. In \href{https://sites.google.com/view/pocr}{Supplementary Materials} II-F, we ablate the number of demonstrations and show that \methodname continues to improve with more data whereas the baselines tend to plateau.

\subsection{Real-World Experiments}
\label{sec:real-env-method}

\textbf{Environment.} To test our algorithm on real-world robotic manipulation tasks, we design an environment 
that consists of a counter-top kitchen setup, in which a Franka robot is tasked with placing various fruits, apple, eggplant, and pineapple
in the green pot located on the far side of the table. Numerous distractors (e.g., toaster, black pot, black pan, burger plate) are placed on the table to create a more visually realistic kitchen countertop scene, bringing the total number of objects to 10.  We use a single 3rd-person monocular RGB camera for policy learning (see Figure~\ref{fig:realworld} for the camera view). For each trained policy, we run 10 trials per task, randomizing the positions of all fruit objects, and we use the identical set of object randomizations for all policies. 

\begin{figure*}[!htbp]
\centering 
% \vspace{-0.2in}
\includegraphics[width=1.0\textwidth]{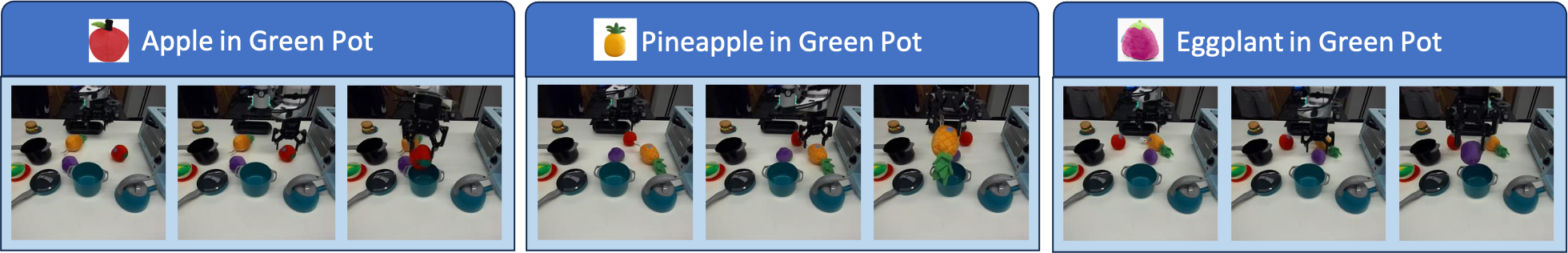}
\vspace{-0.2in}
\caption{Real-World Policy Rollouts.}
\vspace{-0.15in}
\label{figure:real-policy-reel}
\end{figure*}

\textbf{Policy Learning.} 
As in our simulation experiments, we run behavior cloning with keyframe action representation. For each task, we collect 100 trajectories using human teleoperation with the fruits randomly initialized in the center workspace of the table for each trajectory. As it is typical to train visuo-motor control policies in the real world with data augmentation to improve robustness, we train both methods with random cropping augmentation to attain the best performance for all methods; for \methodname, the random-cropping is consistently applied for both the raw RGB input and the masks input. To assess the raw generalization capability of respective representations, we also consider a setup without any data augmentation. 
See \href{https://sites.google.com/view/pocr}{Supplementary Materials} I-B for more experimental details.

\begin{figure}
\centering
% \vspace{-0.15in}
\includegraphics[width=1.0\columnwidth]{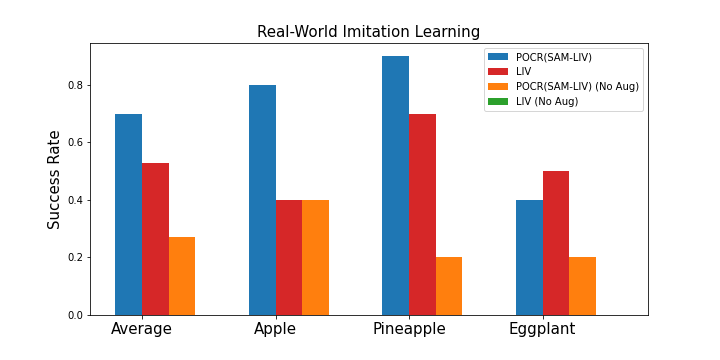}
\vspace{-0.35in}
\caption{Real-World Imitation Learning Results.}
\vspace{-0.2in}
\label{figure:realrobot-main}
% \end{wrapfigure}
\end{figure}

\textbf{Real-World Results.} 
For this real robot experiment, we evaluate the best-performing variant of \methodname in simulation, SAM-LIV. 
See Figure~\ref{figure:realrobot-main} for results and Figure~\ref{figure:real-policy-reel} for rollout visualizations. \methodname(SAM-LIV) once again easily outperforms LIV. On the \texttt{apple} task, it achieves more than double the success rate. When trained without augmentation, LIV overfits the demonstrations and fails completely (0\% success rate). 
Even in this very difficult setting, \methodname(SAM-LIV) still achieves non-trivial performance. 
These results highlight the fragility 
of flat scene-level representations, even when they have been trained on large, diverse human videos. 
Given these models' lack of fine-grained object understanding, it is not surprising that they may struggle in more object-oriented tasks and overfit to just several 
demonstrations in the limited data regime. However, as our experiments suggest, it is not that their representations are not compatible with fine-grained object reasoning, but rather that they have not been given the right input observations -- the very issue that can be mitigated with our chaining approach that augments ``what" foundation models by explicitly providing the ``where" from an off-the-shelf segmentation model. See \href{https://sites.google.com/view/pocr}{the project site} for videos of \methodname(SAM-LIV)'s real-world policy rollouts.

\subsection{Systematic Generalization Experiments}

% \begin{wrapfigure}{hr}{0.5\columnwidth}
\begin{figure}
\vspace{-0.1in}
\centering
\subfigure[New pear]{\label{fig:real_new_distractor}\includegraphics[width=0.3\columnwidth]{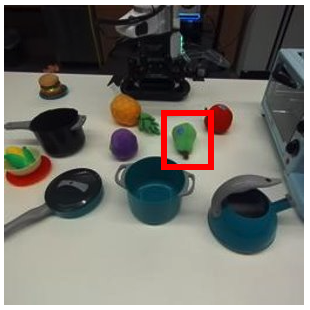}}
\subfigure[New blue cloth]
{\label{fig:real_new_background}\includegraphics[width=0.3\columnwidth]{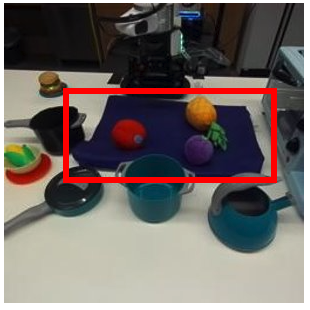}}
\subfigure[New food objects]
{\label{fig:sim_new_distractors}\includegraphics[width=0.3\columnwidth]{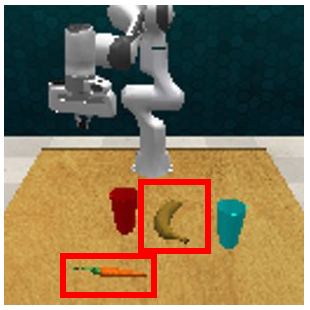}}
\vspace{-0.1in}
\caption{Systematic Generalization Evaluation Environment. Figure~\ref{fig:real_new_distractor}: green pear is the new distractor fruit; Figure~\ref{fig:real_new_background}: blue cloth serves as new background; Figure~\ref{fig:sim_new_distractors} Carrot and banana are new distractors.}
\vspace{-0.2in}
% \end{wrapfigure}
\label{fig:sys_gen_env}
\end{figure}

\textbf{Methods, Training \& Evaluation.} Prior works~\cite{van2019perspective,Greff2020-hq,Lake2016-ze,pmlr-v162-dittadi22a,yoon2023investigation} have established systematic generalization, the model's ability to generalize to unseen scenarios different but semantically similar to the training data, as an advantage of OCEs. To validate the ability of \methodname to achieve systematic generalization, we perform experiments in the real-world and simulated task setting shown in Figure~\ref{fig:sys_gen_env}. In Figure~\ref{fig:real_new_distractor},
the robot must pick up the apple and put it in the green pot, but there is one other ``distractor" fruit, the green pear, in the scene. In Figure~\ref{fig:real_new_distractor}, the robot needs to perform the same task as above, but now in the presence of a new background in the form of a blue cloth. And finally, in Figure~\ref{fig:sim_new_distractors}, the robot must pick up the red cup, but we introduced two unseen distractors, a carrot, and a banana. 
We evaluated the same policies as in the earlier sections in these systematic generalization environments. All new distractor objects were also initialized to random positions.

\textbf{Results.} In every real-world and simulated evaluation setting, \methodname(SAM-LIV) policies are largely unaffected, but the baseline LIV performs significantly worse.  In the real-world setting with the new distractor fruit,  the success rate of \methodname(SAM-LIV) only drops from 70\% to 60\%, whereas LIV drops from 40\% to 20\%. These trends are even more pronounced in the new background setting (also in real-world): again, \methodname(SAM-LIV) is only marginally affected, dropping from 70\% to 50\%, while LIV, unable to adapt to this out-of-distribution scenario, drops to 0\%. 
In our experiments, the LIV policy, unable to handle the out-of-distribution visual observations, consistently performs meaningless and dangerous actions never seen in training, hyper-extending the robot arm, and nearly causing damage to the experimental setup. \methodname(SAM-LIV) degrades much more gracefully.
Likewise, in the simulated settings with two new distractor objects, \methodname(SAM-LIV) performance decreases only marginally from $79.3 \pm 0.7$ to $76.0 \pm 3.1$. In comparison, the image LIV policy dropped in performance from $49.4 \pm 1.9$ to $27.0 \pm 0.6$. 
These results illustrate the natural advantages of \methodname for systematic generalization. 

% \begin{wrapfigure}{r}{0.45\linewidth}

\section{Limitation and Future Works}
\label{sec:limitations}
\methodname policy inference speed is slow due to using the large, slow SAM model. In settings that require high-frequency policy actions, faster segmentation methods that trade off speed with accuracy may be necessary for practical utility. Our procedure for localizing and assigning objects to slots (Section~\ref{sec:pocr-the-where}) has a set of parameters. In our experiments, we use a unified set of parameters that we believe are generally applicable. But in an arbitrary new environment, additional tuning may be required. Furthermore, we follow prior works on pre-trained robotics representations in evaluating \methodname in tasks that largely have the same set of relevant objects in all train and test episodes. Testing for generalization over different manipulated objects, alongside our systematic generalization experiments above, may provide more thorough insights into the quality of our method and baselines. 

\section{Conclusion}
\label{sec:conclusion}
We have presented a simple yet effective recipe for generating reusable object-centric representations for visual robotic manipulation from plugging and playing ``what'' and ``where'' foundation models.
Our framework uses off-the-shelf instance segmentation to produce high-quality, self-consistent masks over time and uses off-the-shelf visual representations to acquire fine-grained visual descriptors for policy learning. Instantiated using state-of-art visual foundation models, \methodname substantially outperforms the state-of-art representations in both simulation and the real world without any in-domain object-centric representation learning. Since POCR can flexibly plug and play with any ``what'' and ``where'' visual foundation models, it has the potential to distill knowledge from more diverse datasets through better pre-trained models in the future.


%%%%%%%%%%%%%%%%%%%%%%%%%%%%%%%%%%%%%%%%%%%%%%%%%%%%%%%%%%%%%%%%%%%%%%%%%%%%%%%%

%%%%%%%%%%%%%%%%%%%%%%%%%%%%%%%%%%%%%%%%%%%%%%%%%%%%%%%%%%%%%%%%%%%%%%%%%%%%%%%%



\section*{ACKNOWLEDGMENT}
We thank Chenxi Dong for assistance with setting up expanded experiments for the final paper and colleagues at UPenn for their insightful feedback. This work is funded by NSF CAREER Award 2239301, ONR award N00014-22-1-2677, a gift from AWS AI for research in Trustworthy AI, and a Penn University Research Foundation (URF) award.

% The preferred spelling of the word ÒacknowledgmentÓ in America is without an ÒeÓ after the ÒgÓ. Avoid the stilted expression, ÒOne of us (R. B. G.) thanks . . .Ó  Instead, try ÒR. B. G. thanksÓ. Put sponsor acknowledgments in the unnumbered footnote on the first page.

%%%%%%%%%%%%%%%%%%%%%%%%%%%%%%%%%%%%%%%%%%%%%%%%%%%%%%%%%%%%%%%%%%%%%%%%%%%%%%%%

% References are important to the reader; therefore, each citation must be complete and correct. If at all possible, references should be commonly available publications.

\clearpage
% \bibliographystyle{IEEEtran}
% \bibliography{IEEEabrv,IEEEexample}

% \nocite{*}
% \bibliographystyle{plainnat}
% \bibliography{references}
% \printbibliography[references]
\printbibliography

%%%% ICRA does not allow appendix %%%%
% \clearpage
% \appendices

\clearpage
\appendices

\section{Experimental details}
\subsection{Simulation Experimental Details}\label{app:sim-exp-details}
\textbf{Keyframe action representation.} Following the setup of \cite{james2022q}, we perform keyframe discovery on the demonstration dataset to reduce the task horizon. Iterating over each of the demo trajectories, we use a Boolean function to decide whether each trajectory point is a keyframe. The Boolean function is a disjunction of change in the gripper state and velocities approaching near zero. 

\subsection{Real Robot Experimental Details}\label{app:real-exp-details}
\begin{figure}[ht]
    \centering
    \includegraphics[width=0.8\columnwidth]{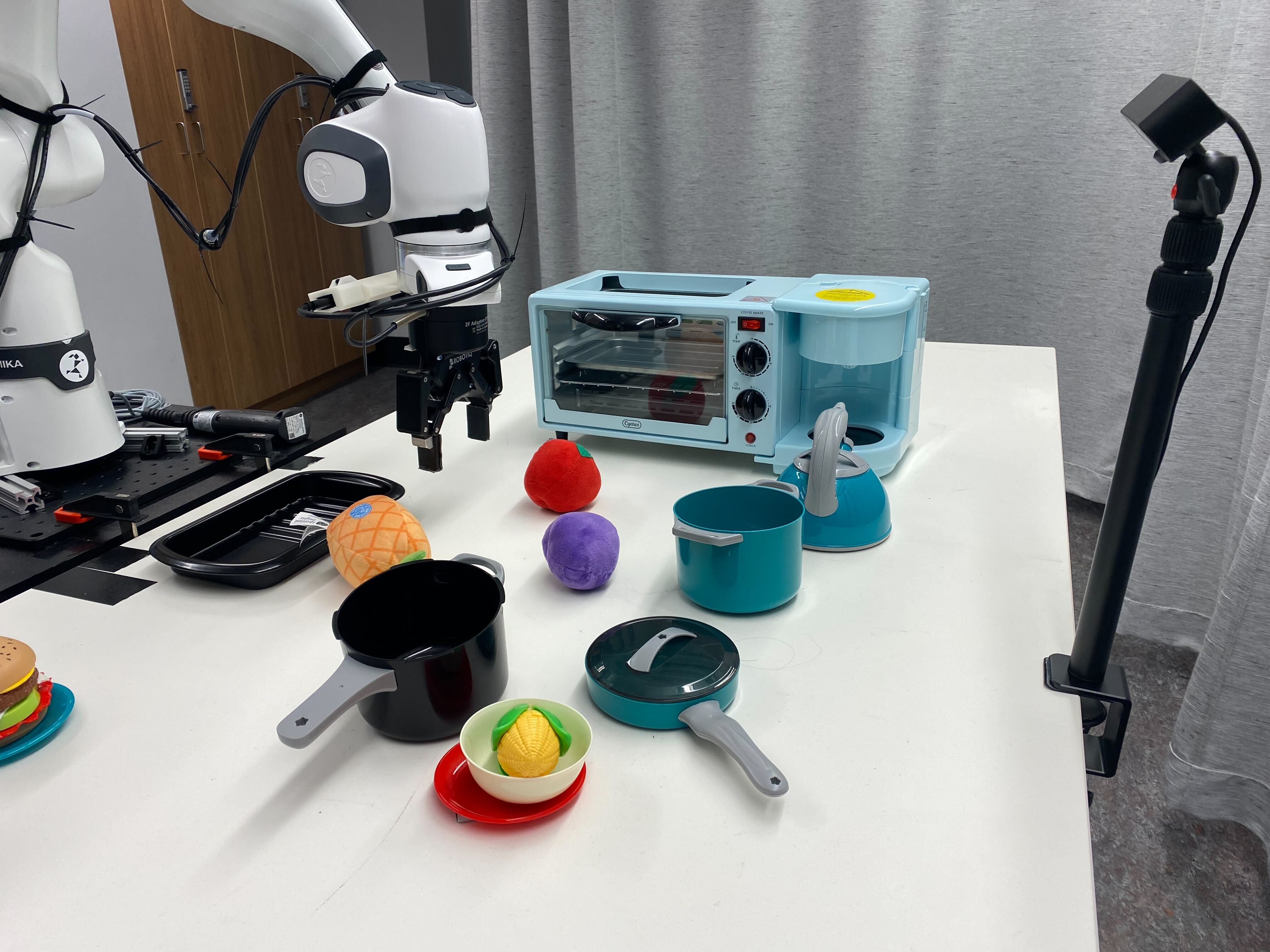}
    \caption{RealRobot environment side view.}
    \label{fig:realworld-side}
\end{figure}

\textbf{RealRobot Environment.} In Figure~\ref{fig:realworld-side}, we show a side view of the RealRobot environment to better illustrate the position of the camera that is used for policy learning.

\revise{\textbf{Robot Mask.} We find empirically that SAM sometimes struggles with consistent segmentation of the robot object in RealRobot environment. Therefore, we remove the robot from our object binding procedure (see Section III-A)
automatically with no manual intervention following procedures in prior works~\cite{hu2021know, bahl2022human}.}

\subsection{Imitation Learning Hyperparameters}
Table~\ref{table:bc-hyperparameters} lists the imitation learning hyperparameters for both RLBench simulation environment and RealRobot environment.
\begin{table}[ht]
\centering
\caption{Imitation Learning Hyperparameters.}\label{table:bc-hyperparameters}
\resizebox{.5\textwidth}{!}{% <------ Don't forget this %
\begin{tabularx}{1.2\columnwidth}{p{3.3cm}|p{3.0cm}p{3.0cm}@{}}
\toprule
Hyperparameters & RLBench & RealRobot\\
\midrule 
Self-Attention & N/A & 4 Heads, 256 Hidden Dimension\\ 
MLP Architecture & [256, 256] & [256, 256]\\ 
Non-Linear Activation & Leaky ReLU & ReLU\\ 
\midrule
Optimizer & Adam & Adam\\
Gradient Steps &  250000 & 10000\\ 
Batch Size &  128 & 64 \\ 
Learning Rate & 0.0005 & 0.001\\ 
Augmentation & Demo Augment \cite{james2022q} &  Random Cropping \\ 
\bottomrule
\end{tabularx}
}
\end{table}

\subsection{Complexity of Localizing and Assigning Objects to Slots}\label{app:complexity}
\revise{In Section III-A,
we described our procedure for localizing and assigning objects to slots.} To prepare our method for a new environment, the only step is to train the k-means foreground screening algorithm \cite{Amir2021DeepVF} with 50 images from the demo dataset, which takes 50 seconds. Note that this only needs to be done once upfront in every environment, after which the agent is ready for policy learning.

Below we list the wall-clock time of each component of our
policy when deployed after training: 
\begin{itemize}
    \item Non-maximum suppression: 0.0027s
    \item SAM mask generation: 1.5s
    \item Hungarian matching: 0.0012s
\end{itemize}

So each inference step takes at most 1.6s. Evidently, the main bottleneck is SAM, but it can be potentially improved by substituting SAM with its faster variants such as FastSAM \cite{zhao2023fast}. It is also not a critical issue in our experiments, since our keyframe trajectory representation significantly limits the number of transitions in a trajectory. Note that the keyframe trajectory representation used in RLBench means that we only need to infer representations and policy actions a handful of times in each rollout.

We use the Jonker-Volgenant algorithm \cite{JValgo} for our Hungarian matching. The complexity is $O(n^3)$ with regard to number of objects. In our environments, we typically have fewer than 20 objects, so Hungarian matching is very fast.

\section{Addition Experimental Results}
\subsection{\revise{Investigating Explicit ``Where'' Representations for \methodname}}\label{app:explicit-mask-rep}
\revise{In Section III-A,
we stated that the “where” component of \methodname is explicitly represented by the axis-aligned bounding box coordinates of each mask. We ablate this choice by comparing two explicit methods of representing masks, bounding box and centroid coordinates. We compare them in two settings: with LIV as the ``what'' representation (\textbf{SAM-LIV}) and without additional ``what" representations (\textbf{SAM}). As shown in Table~\ref{table:bbox-vs-centroid}, since bounding box coordinates result in slightly better performance than centroid coordinates, we opt to use the bounding box representation throughout our other experiments.}

\begin{table}[ht]
% \begin{wraptable}{r}{0.63\textwidth}
\centering
% \vspace{-0.2cm}
\caption{Behavioral cloning success rates (avg. over 100 rollouts), comparing two explicit mask representations.}
\resizebox{.48\textwidth}{!}{% <------ Don't forget this %
\begin{tabularx}{1.2\columnwidth}{p{3.5cm}@{}|p{3cm}@{}p{3cm}@{}}
\toprule
Representations & \texttt{Pick up Cup} & \texttt{Rubbish in Bin} \\
\midrule
SAM(w/ bbox) & 55.7 $\pm$ 1.8 & 24.6 $\pm$ 1.3 \\
SAM(w/ centroid) & 52.7 $\pm$ 1.7 & 24.0 $\pm$ 3.1 \\
\\
SAM-LIV(w/ bbox) & 79.3 $\pm$ 0.7 & 35.3 $\pm$ 1.8 \\
SAM-LIV(w/ centroid) & 74.0 $\pm$ 0.0 & 32.3 $\pm$ 2.4 \\
\bottomrule
\end{tabularx}
}
\label{table:bbox-vs-centroid}
% \vspace{-0.5cm}
\end{table}

\revise{It is reasonable to ask whether any explicit mask representation is necessary at all, given that the ``what'' representations encode masked images (see Section III-B),
which already implicitly contain ``where'' information of each mask. In Table~\ref{table:what-reps-nobbox}, we compare various ``what'' representations for \methodname with and without mask bounding box (\textbf{-bbox}). For every ``what'' representation, removing the bounding box results in significantly worse performance. These results indicate that the ``where'' information is not well captured in the ``what'' encodings, and we need to explicitly represent them separately. Therefore, throughout our other experiments, we include explicit ``where'' representations.}

\begin{table}[ht]
% \begin{wraptable}{r}{0.63\textwidth}
\centering
% \vspace{-0.2cm}
\caption{Behavioral cloning success rates (avg. over 100 rollouts), comparing various ``What" Representations for \methodname with and without explicit mask representations.}
\resizebox{.48\textwidth}{!}{% <------ Don't forget this %
\begin{tabularx}{1.2\columnwidth}{p{3.5cm}@{}p{3cm}@{}p{3cm}@{}}
\toprule
Representations & \texttt{Pick up Cup} & \texttt{Rubbish in Bin} \\
\midrule
SAM-LIV & 79.3 $\pm$ 0.7 & 35.3 $\pm$ 1.8 \\
SAM-LIV(-bbox) & 63.7 $\pm$ 0.3 & 25.0 $\pm$ 1.0 \\
\\
SAM-R3M & 78.0 $\pm$ 1.3 & 33.3 $\pm$ 1.3 \\
SAM-R3M(-bbox) & 55.0 $\pm$ 1.0 & 19.0 $\pm$ 1.7 \\
\\
SAM-ImageNet & 70.7 $\pm$ 0.7 & 30.0 $\pm$ 2.0 \\
SAM-ImageNet(-bbox) & 22.7 $\pm$ 1.3 & 10.0 $\pm$ 1.2 \\
% \midrule
\\
SAM-Scratch & 32.7 $\pm$ 2.9  & 12.0 $\pm$ 2.0 \\
SAM-Scratch(-bbox) & 19.2 $\pm$ 2.5 & 4.4 $\pm$ 0.7 \\
\bottomrule
\end{tabularx}
}
\label{table:what-reps-nobbox}
% \vspace{-0.5cm}
\end{table}

\revise{Comparing the performances of these encoders across the two tasks 
offers interesting insights into the kinds of image information that manipulation policies require. 
SAM(w/ bbox) and SAM(w/ centroid) discard important information about the located objects, e.g. their precise poses, geometries, articulations, textures, and category identities. 
By including additional slot vector representation as the ``what'' component of \methodname, SAM-LIV, SAM-R3M, and SAM-ImageNet perform significantly better than using simple mask shape properties. Additionally, since the object binding procedure described in Section III-A
has an overall mask assignment accuracy of 94.4\% (see Appendix~\ref{app:obj-binding} for more details), SAM(w/ bbox) struggles to account for noise in mask assignments without additional information provided by slot vector representation.}

\subsection{\revise{Comparing Performance using SAM and Ground Truth Masks}}\label{app:sam-vs-gt}
\revise{We compare \methodname's downstream control performance using SAM masks (\textbf{SAM}) and ground truth masks (\textbf{GT}) provided by the simulation to study the gap between SAM and an upper-bound baseline. As shown in Table~\ref{table:sam-vs-gt}, there exists a small gap between SAM masks and ground truth masks in terms of their downstream control performance. This demonstrates that SAM can generate masks that almost match ground truth masks in quality, but there still exists a small amount of inaccuracies and noise.}

\begin{table}[ht]
% \begin{wraptable}{r}{0.63\textwidth}
\centering
% \vspace{-0.2cm}
\caption{Behavioral cloning success rates (avg. over 100 rollouts), comparing SAM and ground truth masks.}
\resizebox{.48\textwidth}{!}{% <------ Don't forget this %
\begin{tabularx}{1.2\columnwidth}{p{3.5cm}@{}|p{3cm}@{}p{3cm}@{}}
\toprule
Representations & \texttt{Pick up Cup} & \texttt{Rubbish in Bin} \\
\midrule
GT-LIV & 82.7 $\pm$ 2.4 & 38.3 $\pm$ 1.8  \\
SAM-LIV & 79.3 $\pm$ 0.7 & 35.3 $\pm$ 1.8  \\
\bottomrule
\end{tabularx}
}
\label{table:sam-vs-gt}
% \vspace{-0.5cm}
\end{table}

\subsection{\revise{Comparing \methodname to Baselines using BC-RNN framework}}\label{app:bc-rnn}
\revise{BC-RNN is a popular approach for recurrent imitation policy learning used by prior works~\cite{robomimic2021, Zhu2022VIOLA}. We follow the BC-RNN implementation and hyperparameters of RoboMimic~\cite{robomimic2021}. We explore four variants of BC-RNN: BC-RNN integrated with POCR (SAM-LIV), LIV, R3M, and ResNet-50 pre-trained on the ImageNet dataset as visual representations. As shown in Table~\ref{table:bc-rnn}, all variants fall far short of POCR (SAM-LIV) in the BC-RNN setting, similar to our conclusion in the standard BC setting (Section V-D).
}

\begin{table}[ht]
% \begin{wraptable}{r}{0.63\textwidth}
\centering
% \vspace{-0.2cm}
\caption{Behavioral cloning success rates (avg. over 100 rollouts), comparing various representations for BC-RNN.}
\resizebox{.48\textwidth}{!}{% <------ Don't forget this %
\begin{tabularx}{1.2\columnwidth}{p{5cm}|p{5cm}}
\toprule
Representations & \texttt{Pick up Cup}  \\
\midrule
POCR (SAM-LIV) & 78.7 $\pm$ 2.4 \\
LIV & 52.5 $\pm$ 2.4 \\
R3M & 56.0 $\pm$ 3.1 \\
ImageNet & 14.5 $\pm$ 0.8 \\
\bottomrule
\end{tabularx}
}
\label{table:bc-rnn}
% \vspace{-0.5cm}
\end{table}

% \begin{table*}[!htbp]
% % \begin{wraptable}{r}{0.63\textwidth}
% \caption{\methodname Object Binding Accuracy}
% % \vspace{-0.1in}
% \centering
% \resizebox{1.0\textwidth}{!}{% <------ Don't forget this %
% \begin{tabularx}{1.1\textwidth}{c
% |{2.5cm}c{2.5cm}c{2.5cm}c{2.5cm}|c{2.5cm}c{2.5cm}c{2.5cm}c{2.5cm}|c{2.5cm}}
% \toprule
%  Task & Pick up Cup & Rubbish in Bin & Stack Wine & Phone on Base & Water Plants & Close Box & Close Laptop & \textbf{Overall} \\
% \midrule
% Accuracy & 91.7\% & 87.2\% & 80.5\% & 76.5\% & 98.6\% & 92.6\% & 99.4\% & \textbf{94.4\%} \\
% \bottomrule
% \end{tabularx}
% }
% \label{table:object-binding-accuracy}
% % \vspace{-0.1in}
% \end{table*}

% \begin{table*}[!htbp]
% % \begin{wraptable}{r}{0.63\textwidth}
% \caption{\methodname Object Binding Accuracy}
% % \vspace{-0.1in}
% \centering
% \resizebox{\textwidth}{!}{% <------ Don't forget this %
% \begin{tabularx}{1.1\textwidth}{m
% |m{2.5cm}m{2.5cm}m{2.5cm}|m{2.5cm}m{2.5cm}m{2.5cm}m{2.5cm}|m{2.5cm}}
% \toprule
%  Task & Pick up Cup & Rubbish in Bin & Stack Wine & Phone on Base & Water Plants & Close Box & Close Laptop & \textbf{Overall} \\
% \midrule
% Accuracy & 91.7\% & 87.2\% & 80.5\% & 76.5\% & 98.6\% & 92.6\% & 99.4\% & \textbf{94.4\%} \\
% \bottomrule
% \end{tabularx}
% }
% \label{table:object-binding-accuracy}
% % \vspace{-0.1in}
% \end{table*}

\begin{table*}[!htbp]
\caption{\methodname Object Binding Accuracy}
\centering
\resizebox{\textwidth}{!}{%
\begin{tabularx}{1.3\textwidth}{X*{9}{>{\centering\arraybackslash}X}}
\toprule
Task & Pick up Cup & Rubbish in Bin & Stack Wine & Phone on Base & Water Plants & Close Box & Close Laptop & \textbf{Overall} \\
\midrule
Accuracy & 91.7\% & 87.2\% & 80.5\% & 76.5\% & 98.6\% & 92.6\% & 99.4\% & \textbf{94.4\%} \\
\bottomrule
\end{tabularx}
}
\label{table:object-binding-accuracy}
\end{table*}

\begin{figure*}[!htbp]
\centering
\includegraphics[width=0.8\textwidth]{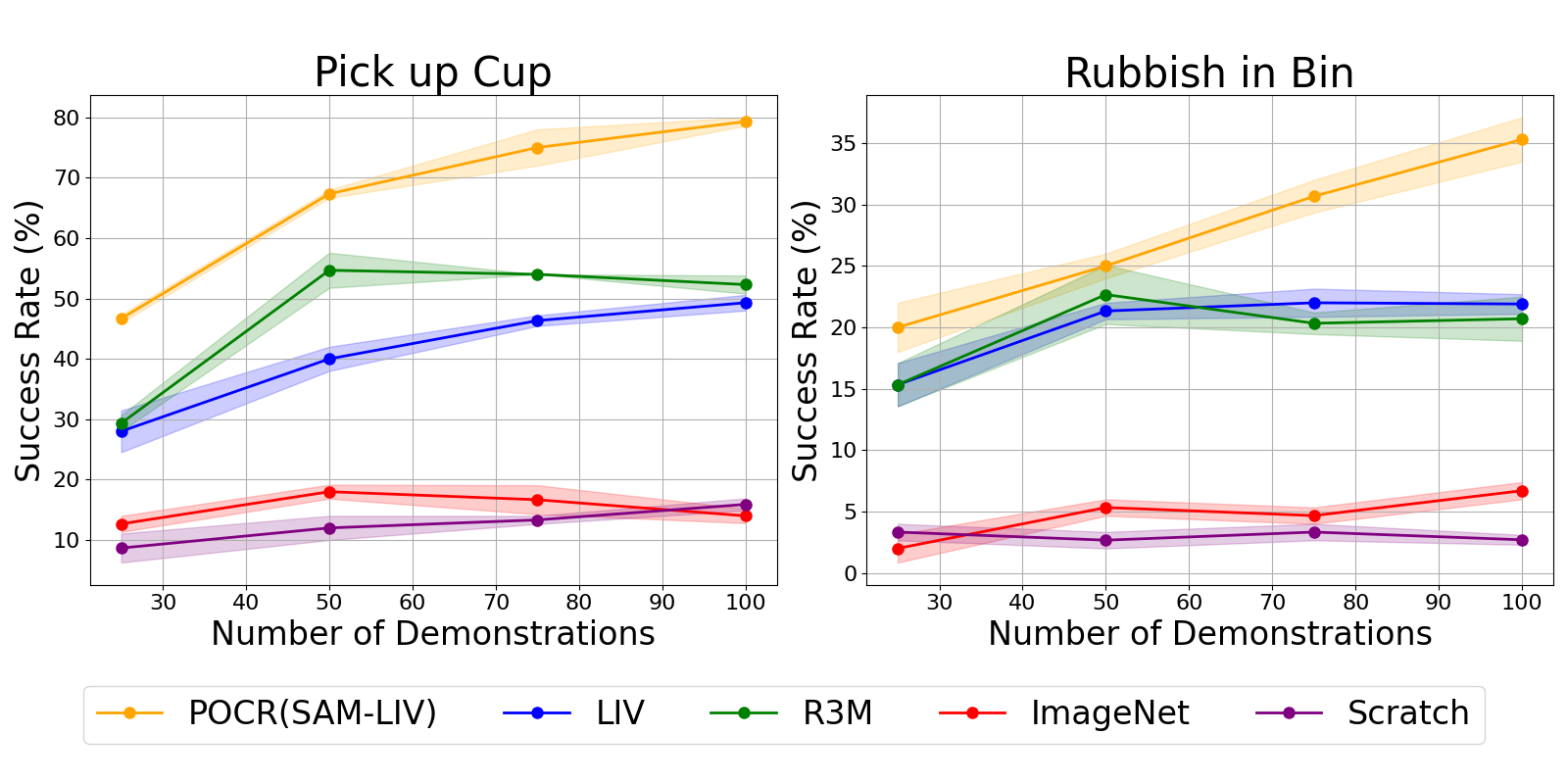}
\caption{\revise{Behavioral cloning success rates over number of demonstrations for each task.
The shaded areas show the standard errors over 3 random seeds. \revisetwo{POCR's performance scales with the dataset size, while the baselines tend to plateau.}}}
\label{figure:ndemos-ablation}
\end{figure*}

\subsection{\revise{Accuracy of \methodname's Object Binding Procedure}\label{app:obj-binding}}
\revise{To better understand the accuracy of POCR's object binding procedure (Section III-A),
we evaluate it quantitatively using the following procedure:}
\revise{\begin{enumerate}
    \item We first run SAM using a reference image to generate segmentation masks, which then act as reference slots for future slot binding.
    \item Using IoU (following the procedure in FG ARI~\cite{Rand1971ObjectiveCF}), ground-truth masks are matched with SAM mask slots to determine the ground-truth slot assignments for all objects in the reference image.
    \item All following frames in the demo dataset are processed with SAM and slot binding to generate \textbf{\methodname's slot assignments}.
    \item We repeat the IoU-based matching procedure between ground-truth masks and SAM mask slots for all frames in the demonstration dataset to obtain the \textbf{ground-truth slot assignments}.
    \item Finally, we compare \methodname's slot assignments (step 3) with the ground-truth slot assignments (step 4) to calculate the accuracy of our object binding procedure. 
\end{enumerate}}

\revise{Table~\ref{table:object-binding-accuracy} shows the quantitative results. \methodname achieves an overall accuracy of 94.4\%.}

\begin{figure}[ht]
    \centering
    \includegraphics[width=1.0\columnwidth]{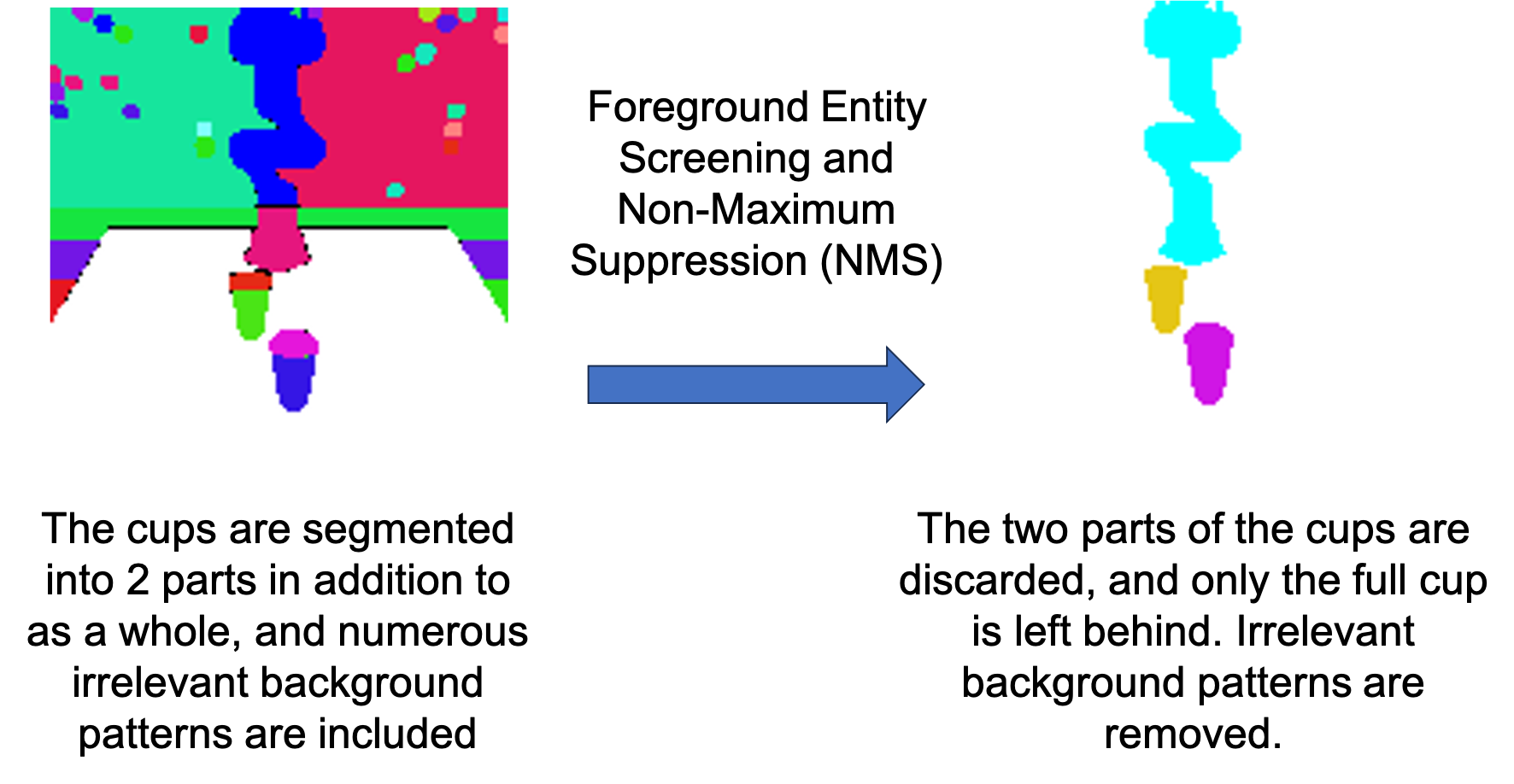}
     \caption{\revisetwo{Automatic removal of unnecessary masks by \methodname. Through screening the object-level foreground entity candidates and non-maximum suppression, we remove irrelevant background and overlapping objects.}}
    \label{fig:foreground-screening}
\end{figure}

\subsection{Foreground Screening and \revisetwo{Non-Maximum Suppression} Ablation Studies}\label{app:foreground-screening}
We remove background \revisetwo{and overlapping} objects automatically (see Section III-A),
to (1) reduce the sizes of our representation, and subsequently, the policy architecture and expert datasets, and (2) achieve invariance to background distractor objects.

In our simulation, the automatically removed background consists of the table and numerous patterned dots. In our real-world environment, it includes wires, people, and other miscellaneous objects. \revisetwo{Additionally, in both simulation and the real world, we remove object subparts at various levels of granularity.} \revise{See Figure~\ref{fig:foreground-screening} for visualization of \revisetwo{unnecessary masks automatically removed by \methodname}.} While vision-based robot manipulation experiments are often performed with a plain background to avoid distractors, we do not carefully design any such clean background. So this process ensures our method’s versatility across noisy and realistic environments. 

We perform ablation experiments following the experimental setups described in Section V-A.
We removed the ``obtaining object segments in a reference image by screening the object-level foreground entity candidates'' part described in Section III-A, 
resulting in a longer list of SAM mask candidates consisting of numerous patterns in the background. \revise{Performance drops significantly from 79.3 ± 0.7 with foreground screening to 33.0 ± 3.1 without foreground screening for \texttt{Pick up Cup} task in simulation.}

\begin{figure*}[!htbp]
\centering
\includegraphics[width=0.8\textwidth]{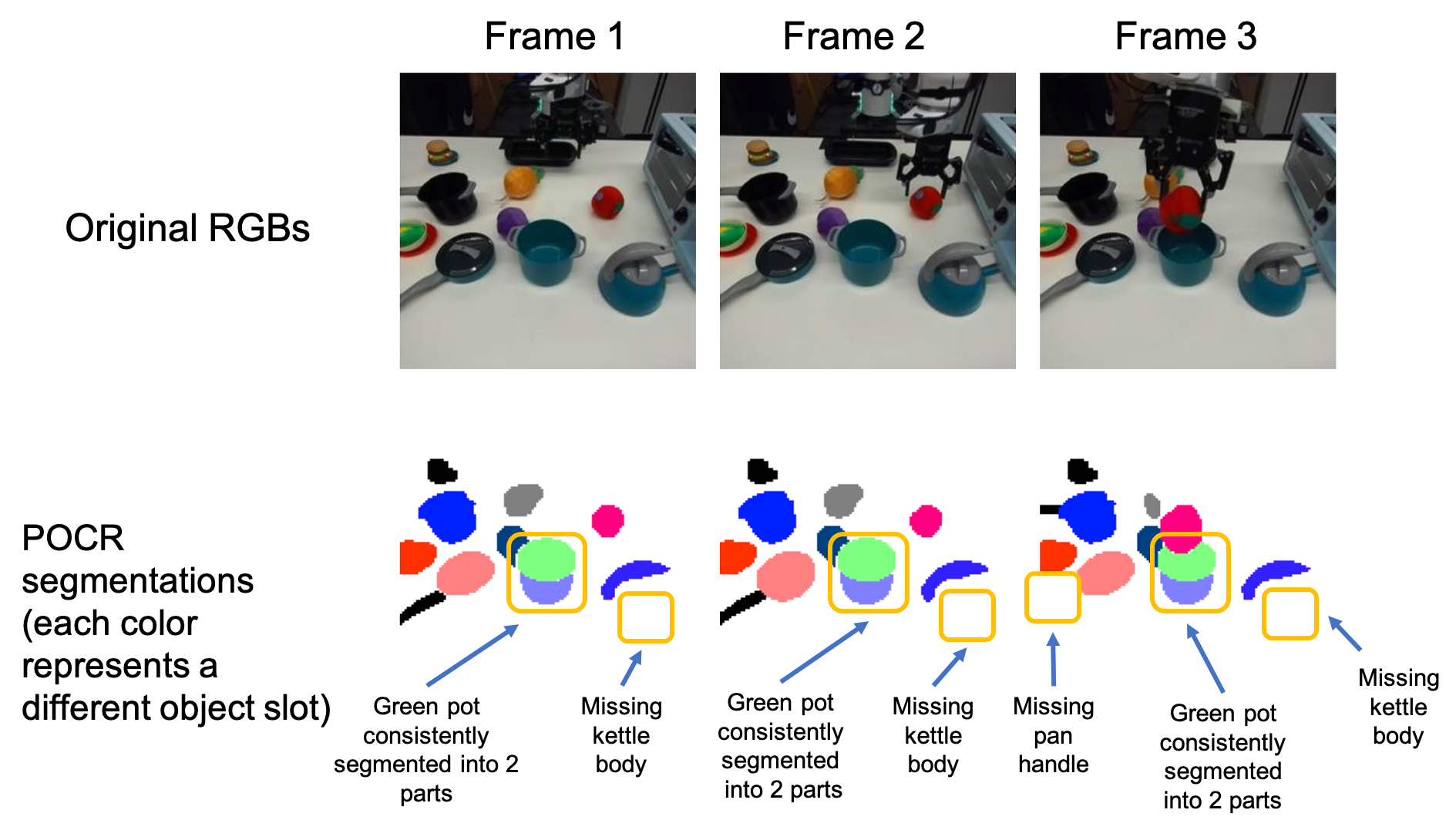}
\caption{\revisetwo{Examples of \methodname segmentation failures. These failures include over-segmentation of an object into its parts and missing certain objects or their parts. As discussed in Appendix~\ref{app:pocr-segment-failures}, barring rare circumstances, these failures would not result in task failures.
}}
\label{figure:segmentation-failures}
\end{figure*}

\subsection{\revise{Number of Demonstrations Ablation Studies}}\label{app:ndemos-ablation}
\revise{We perform ablation studies on the number of demonstrations to understand its effect on \methodname compared to the baselines. For this experiment, we use \texttt{Pick up Cup} and \texttt{Rubbish in Bin}, two simulation tasks in RLBench, and we follow the experimental setups for behavioral cloning described in Section V-A. As shown in Figure~\ref{figure:ndemos-ablation}, across both tasks, \methodname continues to improve as the data size grows. In contrast, LIV grows at a smaller rate than \methodname, and R3M, ImageNet, and Scratch tend to plateau.}

\section{\revise{Additional Qualitative Results}}
\label{app:qualitative}

\subsection{\revise{\methodname Segmentation Failures}}\label{app:pocr-segment-failures}
\revise{While our post-processing procedure helps to clean up SAM masks and reduce errors, it does not fully get rid of them. See Figure~\ref{figure:segmentation-failures} for an example illustration. We find in practice that when the post-processed SAM segments correspond to parts rather than full objects, they are most often consistent between frames. For example, the green pot in Figure~\ref{figure:segmentation-failures} is consistently segmented into two parts in each frame. This is easily handled in our method: that object effectively uses two slots in our representation rather than just one. As long as we set the maximum slots to be large enough, this is not a problem. A more difficult type of error occurs when SAM is inconsistent between frames: e.g. it might sporadically miss an object or a part. Even in these cases, this is easily handled if the object is not task-relevant such as the panhandle in the bottom left: the object-centric structure of the representation permits the downstream policy to easily learn to ignore irrelevant slots. Note that this is likely to be the most frequent error in a heavily cluttered scene with many distractor objects. If such errors occur on a task-relevant object though, it can cause task failures. Overall, we find that relatively few of our policy failures can be clearly attributed to such irrecoverable SAM failures.} 

\subsection{\revise{\revisetwo{Slot-wise Breakdowns} of \methodname Masks}}\label{app:qualitative-pocr-masks} 
\begin{figure*}[!htbp]
\centering
\begin{tabularx}{\textwidth}{
  >{\centering\arraybackslash}m{0.08\textwidth}
  >{\centering\arraybackslash}X
  >{\centering\arraybackslash}X
  >{\centering\arraybackslash}X
  >{\centering\arraybackslash}X
  >{\centering\arraybackslash}X
  >{\centering\arraybackslash}X
  >{\centering\arraybackslash}X
  >{\centering\arraybackslash}X
  >{\centering\arraybackslash}X
  >{\centering\arraybackslash}X
  >{\centering\arraybackslash}X
  >{\centering\arraybackslash}X
}
& \scriptsize\textbf{Image} & \scriptsize\textbf{Background} & \scriptsize\textbf{Slot 1} & \scriptsize\textbf{Slot 2} & \scriptsize\textbf{Slot 3} & \scriptsize\textbf{Slot 4} & \scriptsize\textbf{Slot 5} & \scriptsize\textbf{Slot 6} & \scriptsize\textbf{Slot 7} & \scriptsize\textbf{Slot 8} & \scriptsize\textbf{Slot 9} & \scriptsize\textbf{Slot 10} \\
\vspace*{5mm} \\
\raisebox{-0.5\height}{\scriptsize\textbf{Frame 1}} & \multicolumn{12}{c}{\includegraphics[width=0.92\textwidth, valign=m]{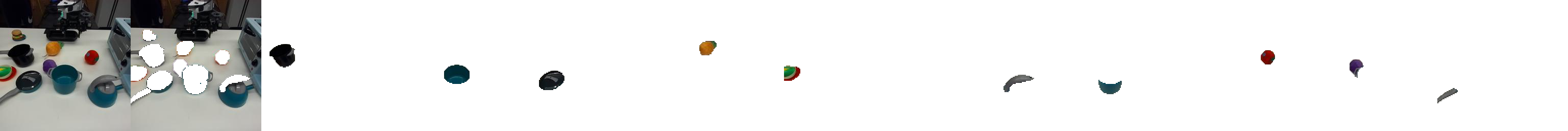}} \\ 
\raisebox{-0.5\height}{\scriptsize\textbf{Frame 2}} & \multicolumn{12}{c}{\includegraphics[width=0.92\textwidth, valign=m]{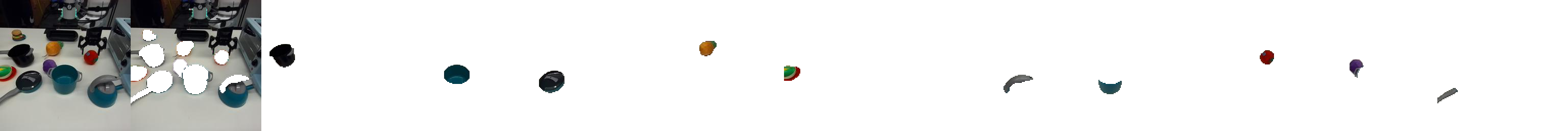}} \\ 
\raisebox{-0.5\height}{\scriptsize\textbf{Frame 3}} & \multicolumn{12}{c}{\includegraphics[width=0.92\textwidth, valign=m]
{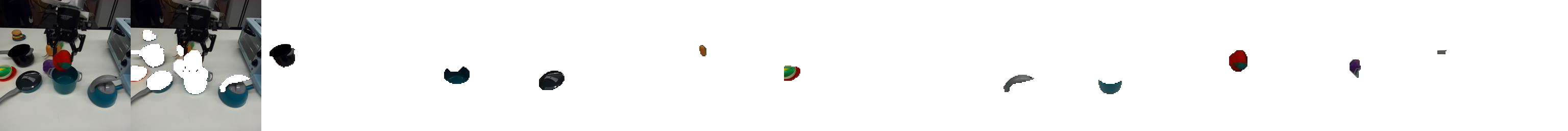}} \\
\end{tabularx}
\vspace*{5mm}
\caption{\revisetwo{Slot-wise masks of policy rollout from RealRobot task \texttt{Apple in Green Pot}. \methodname consistently assigns the target object, the apple, to slot 8. It also correctly assigns a number of distractor objects in the scene, while missing the panhandle in slot 10 of frame 3.}}
\label{figure:apple_separate}
\end{figure*}

\begin{figure*}[!htbp]
\centering
\begin{tabularx}{\textwidth}{
  >{\centering\arraybackslash}m{0.08\textwidth}
  >{\centering\arraybackslash}X
  >{\centering\arraybackslash}X
  >{\centering\arraybackslash}X
  >{\centering\arraybackslash}X
  >{\centering\arraybackslash}X
  >{\centering\arraybackslash}X
  >{\centering\arraybackslash}X
  >{\centering\arraybackslash}X
  >{\centering\arraybackslash}X
  >{\centering\arraybackslash}X
  >{\centering\arraybackslash}X
  >{\centering\arraybackslash}X
}
& \scriptsize\textbf{Image} & \scriptsize\textbf{Background} & \scriptsize\textbf{Slot 1} & \scriptsize\textbf{Slot 2} & \scriptsize\textbf{Slot 3} & \scriptsize\textbf{Slot 4} & \scriptsize\textbf{Slot 5} & \scriptsize\textbf{Slot 6} & \scriptsize\textbf{Slot 7} & \scriptsize\textbf{Slot 8} & \scriptsize\textbf{Slot 9} & \scriptsize\textbf{Slot 10} \\
\vspace*{5mm} \\
\raisebox{-0.5\height}{\scriptsize\textbf{Frame 1}} & \multicolumn{12}{c}{\includegraphics[width=0.92\textwidth, valign=m]{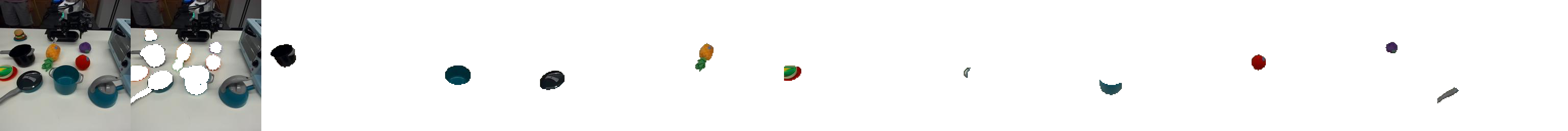}} \\ 
\raisebox{-0.5\height}{\scriptsize\textbf{Frame 2}} & \multicolumn{12}{c}{\includegraphics[width=0.92\textwidth, valign=m]{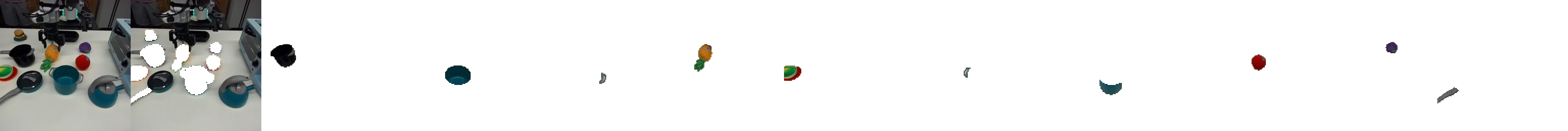}} \\ 
\raisebox{-0.5\height}{\scriptsize\textbf{Frame 3}} & \multicolumn{12}{c}{\includegraphics[width=0.92\textwidth, valign=m]{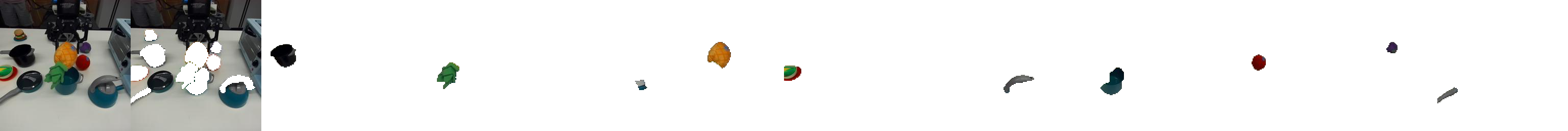}} \\
\end{tabularx}
\vspace*{5mm}
\caption{\revisetwo{Slot-wise masks of policy rollout from RealRobot task \texttt{Pineapple in Green Pot}. \methodname consistently assigns the target object, the pineapple, to slot 4. It also correctly assigns a number of distractor objects in the scene, while making incorrect assignments of distractor objects in slots 3 and 6.}}
\label{figure:pineapple_separate}
\end{figure*}

\begin{figure*}[!htbp]
\centering
\begin{tabularx}{\textwidth}{
  >{\centering\arraybackslash}m{0.08\textwidth}
  >{\centering\arraybackslash}X
  >{\centering\arraybackslash}X
  >{\centering\arraybackslash}X
  >{\centering\arraybackslash}X
  >{\centering\arraybackslash}X
  >{\centering\arraybackslash}X
  >{\centering\arraybackslash}X
  >{\centering\arraybackslash}X
  >{\centering\arraybackslash}X
  >{\centering\arraybackslash}X
  >{\centering\arraybackslash}X
  >{\centering\arraybackslash}X
}
& \scriptsize\textbf{Image} & \scriptsize\textbf{Background} & \scriptsize\textbf{Slot 1} & \scriptsize\textbf{Slot 2} & \scriptsize\textbf{Slot 3} & \scriptsize\textbf{Slot 4} & \scriptsize\textbf{Slot 5} & \scriptsize\textbf{Slot 6} & \scriptsize\textbf{Slot 7} & \scriptsize\textbf{Slot 8} & \scriptsize\textbf{Slot 9} & \scriptsize\textbf{Slot 10} \\
\vspace*{5mm} \\
\raisebox{-0.5\height}{\scriptsize\textbf{Frame 1}} & \multicolumn{12}{c}{\includegraphics[width=0.92\textwidth, valign=m]{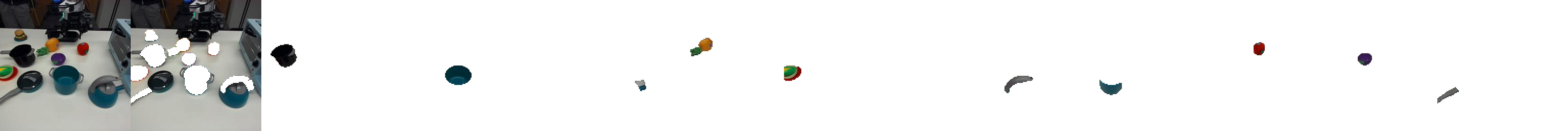}} \\ 
\raisebox{-0.5\height}{\scriptsize\textbf{Frame 2}} & \multicolumn{12}{c}{\includegraphics[width=0.92\textwidth, valign=m]{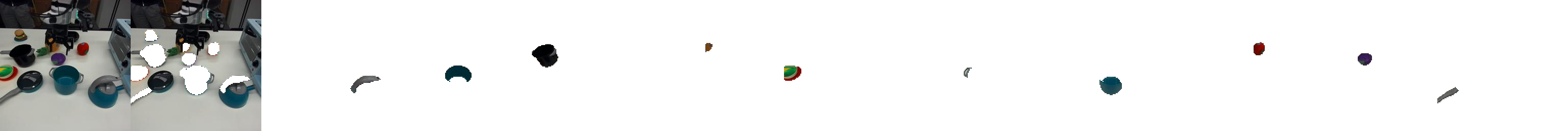}} \\ 
\raisebox{-0.5\height}{\scriptsize\textbf{Frame 3}} & \multicolumn{12}{c}{\includegraphics[width=0.92\textwidth, valign=m]{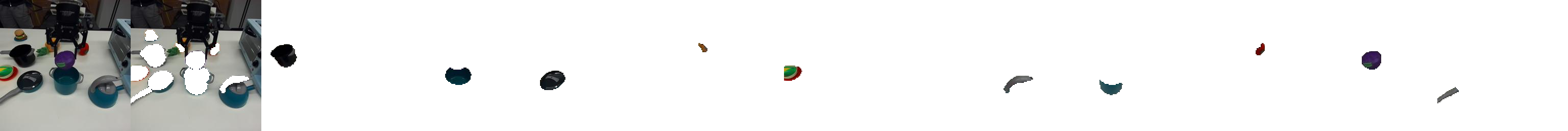}} \\
\end{tabularx}
\vspace*{5mm}
\caption{\revisetwo{Slot-wise masks of policy rollout from RealRobot task \texttt{Eggplant in Green Pot}. \methodname consistently assigns the target object, the eggplant, to slot 9. It also correctly assigns a number of distractor objects in the scene, while making incorrect assignments of distractor objects in slots 1, 3, and 6.}}
\label{figure:eggplant_separate}
\end{figure*}

\begin{figure*}[!htbp]
\centering
\begin{tabularx}{\textwidth}{
  >{\centering\arraybackslash}m{0.16\textwidth}
  >{\centering\arraybackslash}X
  >{\centering\arraybackslash}X
  >{\centering\arraybackslash}X
  >{\centering\arraybackslash}X
  >{\centering\arraybackslash}X
}
& \textbf{Image} & \textbf{Background} & \textbf{Slot 1} & \textbf{Slot 2} & \textbf{Slot 3} \\
\vspace*{5mm} \\
\raisebox{0.5\height}{\textbf{Frame 1}} & \multicolumn{5}{c}{\includegraphics[width=0.83\textwidth, valign=m]{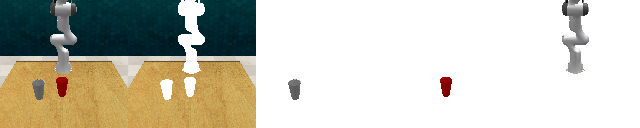}} \\ 
\raisebox{0.5\height}{\textbf{Frame 2}} & \multicolumn{5}{c}{\includegraphics[width=0.83\textwidth, valign=m]{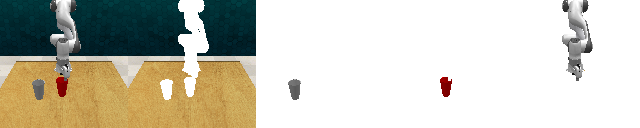}} \\ 
\raisebox{0.5\height}{\textbf{Frame 3}} & \multicolumn{5}{c}{\includegraphics[width=0.83\textwidth, valign=m]{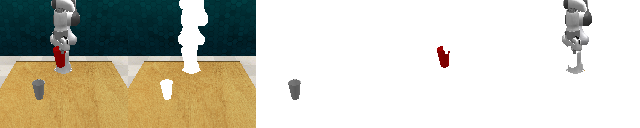}}\\
\end{tabularx}
\vspace*{5mm}
\caption{\revisetwo{Slot-wise masks of \methodname policy rollout from RLBench task \texttt{Pick up Cup}. \methodname consistently assigns the grey distractor cup, the red target cup, and the robot to their respective slots throughout the policy rollout episode.}}
\label{figure:cup_separate}
\end{figure*}

\begin{figure*}[!htbp]
\centering
\begin{tabularx}{\textwidth}{
  >{\centering\arraybackslash}m{0.125\textwidth}
  >{\centering\arraybackslash}X
  >{\centering\arraybackslash}X
  >{\centering\arraybackslash}X
  >{\centering\arraybackslash}X
  >{\centering\arraybackslash}X
  >{\centering\arraybackslash}X
  >{\centering\arraybackslash}X
}
& \textbf{Image} & \textbf{Background} & \textbf{Slot 1} & \textbf{Slot 2} & \textbf{Slot 3} & \textbf{Slot 4} & \textbf{Slot 5} \\
\vspace*{5mm} \\
\raisebox{0.5\height}{\textbf{Frame 1}} & \multicolumn{7}{c}{\includegraphics[width=0.875\textwidth, valign=m]{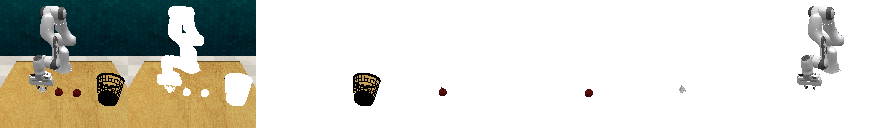}} \\ 
\raisebox{0.5\height}{\textbf{Frame 2}} & \multicolumn{7}{c}{\includegraphics[width=0.875\textwidth, valign=m]{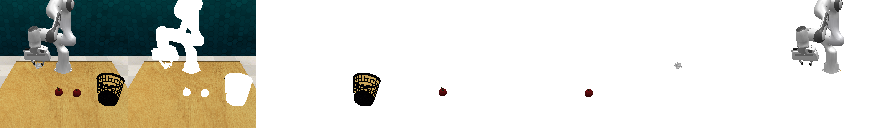}} \\ 
\raisebox{0.5\height}{\textbf{Frame 3}} & \multicolumn{7}{c}{\includegraphics[width=0.875\textwidth, valign=m]{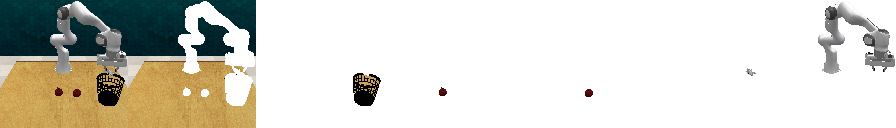}} \\
\end{tabularx}
\vspace*{5mm}
\caption{\revisetwo{Slot-wise masks of \methodname policy rollout from RLBench task \texttt{Rubbish in Bin}. \methodname consistently assigns the rubbish bin, the two distractor apples, the paper rubbish, and the robot to their respective slots throughout the policy rollout episode.}}
\label{figure:bin_separate}
\end{figure*}

\begin{figure*}[!htbp]
\centering
\begin{tabularx}{\textwidth}{
  >{\centering\arraybackslash}m{0.16\textwidth}
  >{\centering\arraybackslash}X
  >{\centering\arraybackslash}X
  >{\centering\arraybackslash}X
  >{\centering\arraybackslash}X
  >{\centering\arraybackslash}X
}
& \textbf{Image} & \textbf{Background} & \textbf{Slot 1} & \textbf{Slot 2} & \textbf{Slot 3} \\
\vspace*{5mm} \\
\raisebox{0.5\height}{\textbf{Frame 1}} & \multicolumn{5}{c}{\includegraphics[width=0.83\textwidth, valign=m]{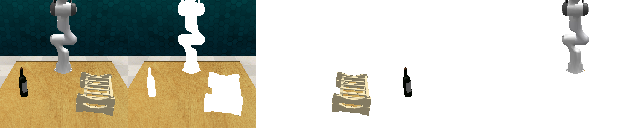}} \\
\raisebox{0.5\height}{\textbf{Frame 2}} & \multicolumn{5}{c}{\includegraphics[width=0.83\textwidth, valign=m]{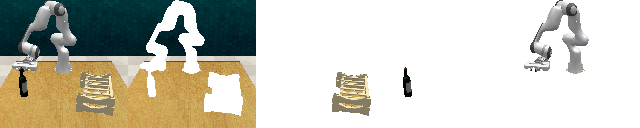}} \\
\raisebox{0.5\height}{\textbf{Frame 3}} & \multicolumn{5}{c}{\includegraphics[width=0.83\textwidth, valign=m]{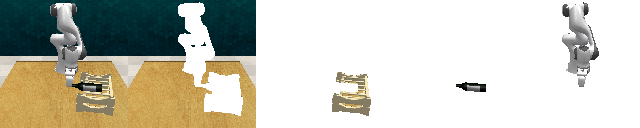}} \\
\end{tabularx}
\vspace*{5mm}
\caption{\revisetwo{Slot-wise masks of \methodname policy rollout from RLBench task \texttt{Stack Wine}. \methodname consistently assigns the wine rack, the wine bottle, and the robot to their respective slots throughout the policy rollout episode.}}
\label{figure:wine_separate}
\end{figure*}

\begin{figure*}[!htbp]
\centering
\begin{tabularx}{\textwidth}{
  >{\centering\arraybackslash}m{0.16\textwidth}
  >{\centering\arraybackslash}X
  >{\centering\arraybackslash}X
  >{\centering\arraybackslash}X
  >{\centering\arraybackslash}X
  >{\centering\arraybackslash}X
}
& \textbf{Image} & \textbf{Background} & \textbf{Slot 1} & \textbf{Slot 2} & \textbf{Slot 3} \\
\vspace*{5mm} \\
\raisebox{0.5\height}{\textbf{Frame 1}} & \multicolumn{5}{c}{\includegraphics[width=0.83\textwidth, valign=m]{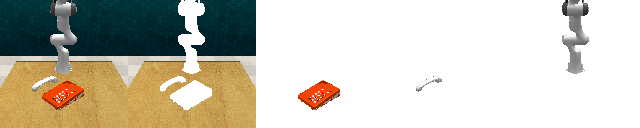}} \\ 
\raisebox{0.5\height}{\textbf{Frame 2}} & \multicolumn{5}{c}{\includegraphics[width=0.83\textwidth, valign=m]{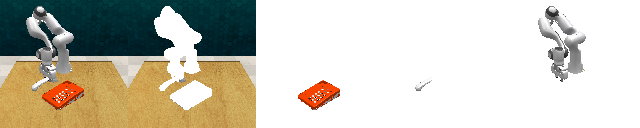}} \\ 
\raisebox{0.5\height}{\textbf{Frame 3}} & \multicolumn{5}{c}{\includegraphics[width=0.83\textwidth, valign=m]{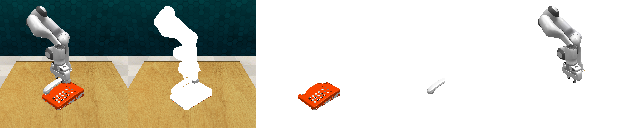}} \\
\end{tabularx}
\vspace*{5mm}
\caption{\revisetwo{Slot-wise masks of \methodname policy rollout from RLBench task \texttt{Phone on Base}. \methodname consistently assigns the phone base, the phone handset, and the robot to their respective slots throughout the policy rollout episode.}}
\label{figure:phone_separate}
\end{figure*}

\begin{figure*}[!htbp]
\centering
\begin{tabularx}{\textwidth}{
  >{\centering\arraybackslash}m{0.142\textwidth}
  >{\centering\arraybackslash}X
  >{\centering\arraybackslash}X
  >{\centering\arraybackslash}X
  >{\centering\arraybackslash}X
  >{\centering\arraybackslash}X
  >{\centering\arraybackslash}X
}
& \textbf{Image} & \textbf{Background} & \textbf{Slot 1} & \textbf{Slot 2} & \textbf{Slot 3} & \textbf{Slot 4} \\
\vspace*{5mm} \\
\raisebox{0.5\height}{\textbf{Frame 1}} & \multicolumn{6}{c}{\includegraphics[width=0.857\textwidth, valign=m]{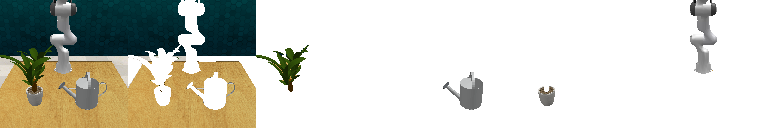}} \\ 
\raisebox{0.5\height}{\textbf{Frame 2}} & \multicolumn{6}{c}{\includegraphics[width=0.857\textwidth, valign=m]{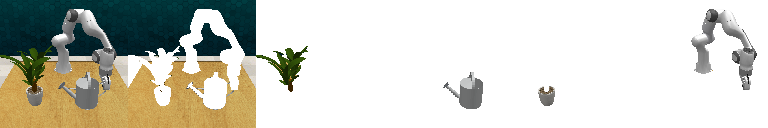}} \\ 
\raisebox{0.5\height}{\textbf{Frame 3}} & \multicolumn{6}{c}{\includegraphics[width=0.857\textwidth, valign=m]{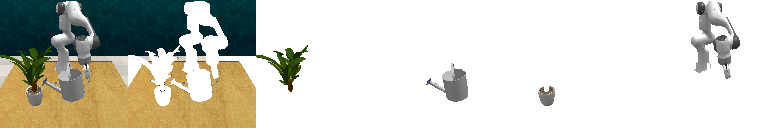}} \\
\end{tabularx}
\vspace*{2mm}
\caption{\revisetwo{Slot-wise masks of \methodname policy rollout from RLBench task \texttt{Water Plants}. \methodname consistently assigns the plant, the watering can, the plant pot, and the robot to their respective slots throughout the policy rollout episode. Note that the top and bottom parts of the plant are separated into two slots. This is not an issue for downstream policy learning as long as they are both tracked consistently.}}
\label{figure:plants_separate}
\end{figure*}

\begin{figure*}[!htbp]
\centering
\begin{tabularx}{\textwidth}{
  >{\centering\arraybackslash}m{0.2\textwidth}
  >{\centering\arraybackslash}X
  >{\centering\arraybackslash}X
  >{\centering\arraybackslash}X
  >{\centering\arraybackslash}X
}
& \textbf{Image} & \textbf{Background} & \textbf{Slot 1} & \textbf{Slot 2} \\
\vspace*{5mm} \\
\raisebox{0.5\height}{\textbf{Frame 1}} & \multicolumn{4}{c}{\includegraphics[width=0.8\textwidth, valign=m]{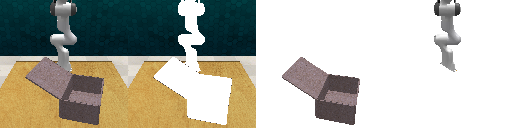}} \\ 
\raisebox{0.5\height}{\textbf{Frame 1}} & \multicolumn{4}{c}{\includegraphics[width=0.8\textwidth, valign=m]{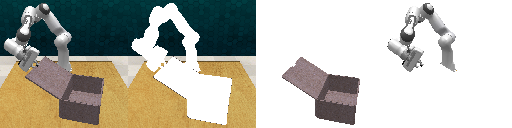}} \\ 
\raisebox{0.5\height}{\textbf{Frame 1}} & \multicolumn{4}{c}{\includegraphics[width=0.8\textwidth, valign=m]{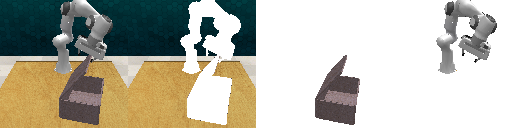}} \\
\end{tabularx}
\vspace*{2mm}
\caption{\revisetwo{Slot-wise masks of \methodname policy rollout from RLBench task \texttt{Close Box}. \methodname consistently assigns the box and the robot to their respective slots throughout the policy rollout episode.}}
\label{figure:box_separate}
\end{figure*}

\begin{figure*}[t]
\centering
\begin{tabularx}{\textwidth}{
  >{\centering\arraybackslash}m{0.2\textwidth}
  >{\centering\arraybackslash}X
  >{\centering\arraybackslash}X
  >{\centering\arraybackslash}X
  >{\centering\arraybackslash}X
}
& \textbf{Image} & \textbf{Background} & \textbf{Slot 1} & \textbf{Slot 2} \\
\vspace*{5mm} \\
\raisebox{0.5\height}{\textbf{Frame 1}} & \multicolumn{4}{c}{\includegraphics[width=0.8\textwidth, valign=m]{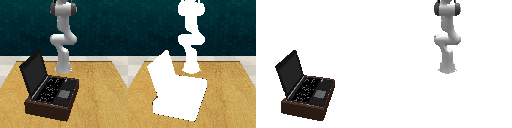}} \\ 
\raisebox{0.5\height}{\textbf{Frame 1}} & \multicolumn{4}{c}{\includegraphics[width=0.8\textwidth, valign=m]{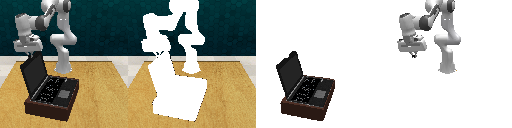}} \\ 
\raisebox{0.5\height}{\textbf{Frame 1}} & \multicolumn{4}{c}{\includegraphics[width=0.8\textwidth, valign=m]{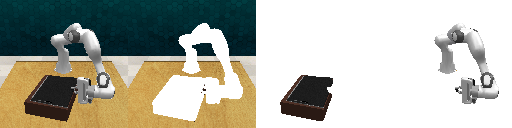}} \\
\end{tabularx}
\vspace*{5mm}
\caption{\revisetwo{Slot-wise masks of \methodname policy rollout from RLBench task \texttt{Close Laptop}. \methodname consistently assigns the laptop and the robot to their respective slots throughout the policy rollout episode.}}
\vspace{100mm}
\label{figure:laptop_separate}
\end{figure*}

\revisetwo{\methodname localizes and assigns objects in each image to object slots (see Section III-A). As mentioned in Section V-B, we visualize slot-wise breakdowns of \methodname masks from each of our real-world and simulation tasks.} \revisetwo{Figures~\ref{figure:apple_separate}-\ref{figure:eggplant_separate} show the slot-wise masks of \methodname policy rollouts from real-world tasks \texttt{Apple in Green Pot}, \texttt{Pineapple in Green Pot}, and \texttt{Eggplant in Green Pot} respectively.}
\revisetwo{Figures~\ref{figure:cup_separate}-\ref{figure:laptop_separate} show the slot-wise masks of \methodname policy rollouts from RLBench tasks \texttt{Pick up Cup}, \texttt{Rubbish in Bin}, \texttt{Stack Wine}, \texttt{Phone on Base}, \texttt{Water Plants}, \texttt{Close Box}, and \texttt{Close Laptop} respectively.}
\revisetwo{These figures demonstrate that \methodname utilizes SAM to provide consistent and accurate tracking of object masks in each slot.
Note that the image and background slots are not used by \methodname during policy learning. They are only included for visualization purposes.} 

\clearpage

\end{document}